%% file: icml2026_main.tex
\documentclass{article}

\input{math_commands.tex}

\usepackage{microtype}
\usepackage{graphicx}
\usepackage{subcaption}
\usepackage{booktabs} %

\usepackage{hyperref}
\usepackage[table]{xcolor} %
\newcommand{\mth}[1]{\textsf{\footnotesize #1}}
\newcommand{\good}[1]{\cellcolor{impGreen}#1}

\usepackage[preprint]{icml2026}

\usepackage{amsmath}
\usepackage{amssymb}
\usepackage{mathtools}
\usepackage{amsthm}
\usepackage[textsize=tiny]{todonotes}
\usepackage{hyperref}
\usepackage{url}
\usepackage{tcolorbox}
\usepackage{graphicx}
\usepackage{array,tabularx,multirow,makecell}

\usepackage{booktabs}   %
\usepackage{multirow}
\usepackage{siunitx}

\usepackage{amsthm}

\usepackage[capitalize,noabbrev]{cleveref}

\theoremstyle{plain}
\newtheorem{theorem}{Theorem}[section]

\theoremstyle{definition}

\theoremstyle{remark}

\renewcommand{\arraystretch}{1.05}      %
\icmltitlerunning{WS-GRPO: Weakly-Supervised Group-Relative Policy Optimization}

\begin{document}

\twocolumn[
  \icmltitle{WS-GRPO: Weakly-Supervised Group-Relative Policy Optimization for Rollout-Efficient Reasoning}

  \icmlsetsymbol{equal}{*}
\renewcommand{\icmlEqualContribution}{*Equal contribution.}
  \begin{icmlauthorlist}
    \icmlauthor{Gagan Mundada}{equal,ucsd}
    \icmlauthor{Zihan Huang}{equal,ucsd}
    \icmlauthor{Rohan Surana}{equal,ucsd}
    \icmlauthor{Sheldon Yu}{ucsd}
    \icmlauthor{Jennifer Yuntong Zhang}{uot}
    \icmlauthor{Xintong Li}{ucsd}
    \icmlauthor{Tong Yu}{adobe}
    \icmlauthor{Lina Yao}{unsw,cicero}
    \icmlauthor{Jingbo Shang}{ucsd}
    \icmlauthor{Julian McAuley}{ucsd}
    \icmlauthor{Junda Wu}{ucsd}
  \end{icmlauthorlist}

  \icmlaffiliation{ucsd}{University of California, San Diego}
  \icmlaffiliation{uot}{University of Toronto}
  \icmlaffiliation{adobe}{Adobe Research}
  \icmlaffiliation{unsw}{The University of New South Wales}
  \icmlaffiliation{cicero}{CSIRO’s Data61}

  \icmlcorrespondingauthor{Junda Wu}{juw069@ucsd.edu}

  \icmlkeywords{Weakly-supervised Learning, GRPO}

  \vskip 0.3in
]

\printAffiliationsAndNotice{\icmlEqualContribution}

\input{contents/0_abstract}

\input{contents/1_introduction}

\input{contents/3_preliminaries}

\input{contents/4_methodology}

\input{contents/5_experiments}

\input{contents/2_related}
\input{contents/6_conclusion}

\section*{Impact Statement}
This paper presents work whose goal is to advance the field of machine learning. There are many potential societal consequences of our work, none of which we feel must be specifically highlighted here.

\input{icml_2026.bbl}
\appendix
\onecolumn

\input{contents/7_appendix}
\end{document}

%% file: math_commands.tex
\usepackage{amsmath,amsfonts,bm}

\def\eqref#1{equation~\ref{#1}}

\def\1{\bm{1}}

\DeclareMathAlphabet{\mathsfit}{\encodingdefault}{\sfdefault}{m}{sl}
\SetMathAlphabet{\mathsfit}{bold}{\encodingdefault}{\sfdefault}{bx}{n}

\DeclareMathOperator*{\argmin}{arg\,min}

%% file: contents/0_abstract.tex
\begin{abstract}
Group Relative Policy Optimization (GRPO) is effective for training language models on complex reasoning. 
However, since the objective is defined relative to a group of sampled trajectories, 
extended deliberation can create more chances to realize relative gains, 
leading to inefficient reasoning and overthinking, and complicating the trade-off between correctness and rollout efficiency. Controlling this behavior is difficult in practice, considering 
(i) Length penalties are hard to calibrate because longer rollouts may reflect harder problems that require longer reasoning, 
penalizing tokens risks truncating useful reasoning along with redundant continuation; 
and (ii) supervision that directly indicates when to continue or stop is typically unavailable beyond final answer correctness.
We propose Weakly Supervised GRPO (WS-GRPO), which improves rollout efficiency by converting terminal rewards into correctness-aware guidance over partial trajectories.
Unlike global length penalties that are hard to calibrate, WS-GRPO trains a preference model from outcome-only correctness to produce prefix-level signals
that indicate when additional continuation is beneficial.
Thus, WS-GRPO supplies outcome-derived continue/stop guidance, reducing redundant deliberation while maintaining accuracy.
We provide theoretical results and empirically show on reasoning benchmarks that WS-GRPO substantially reduces rollout length while remaining competitive with GRPO baselines.
\end{abstract}
\vspace{-20pt}

%% file: contents/1_introduction.tex
\section{Introduction}

\begin{figure}[ht]
    \centering
    \includegraphics[width=1\linewidth]{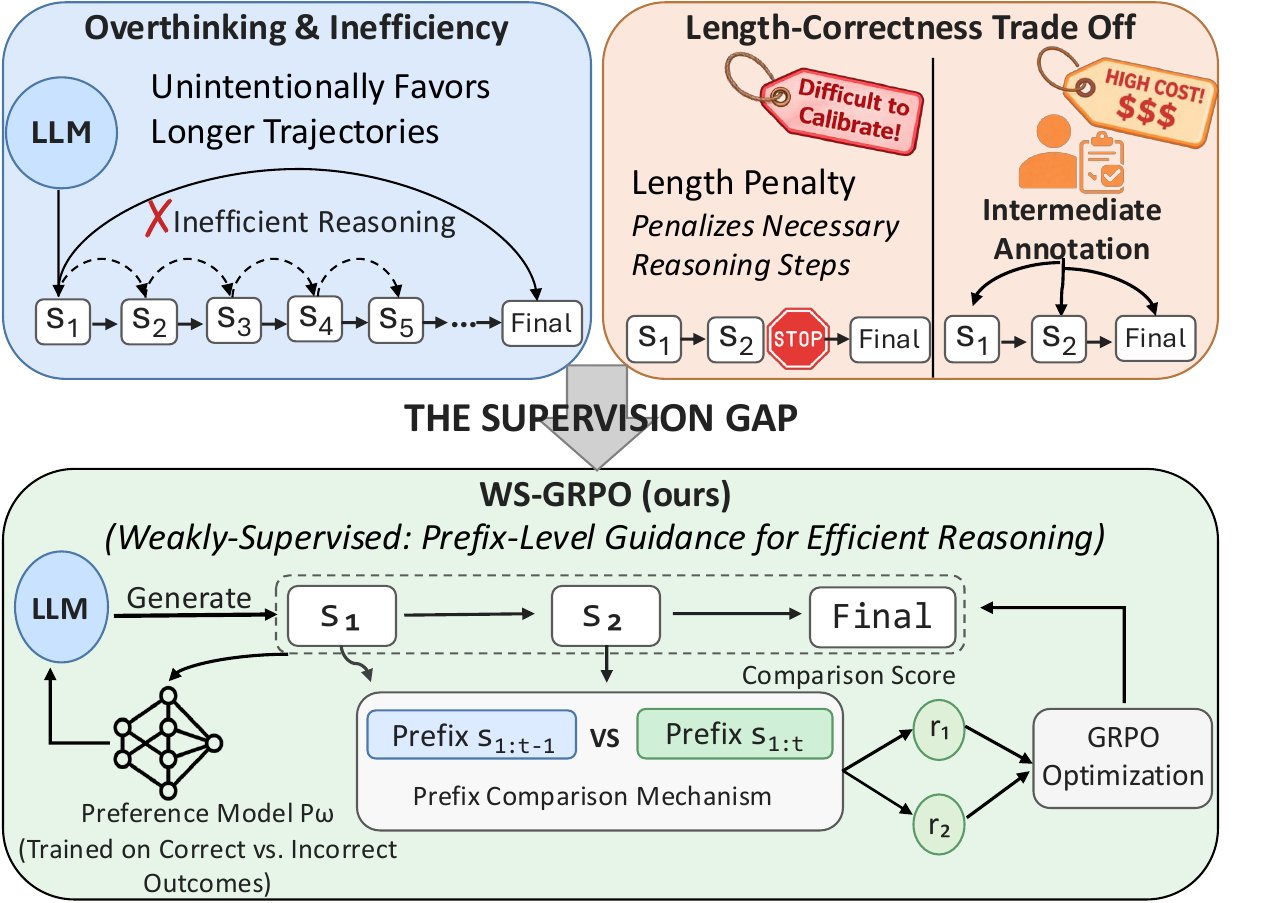}
    \caption{\textbf{WS-GRPO for efficient reasoning.} GRPO's group-relative objective can favor longer trajectories, while simple length penalties are hard to calibrate and human supervision is expensive (top). WS-GRPO uses a preference model trained on correct vs. incorrect outcomes to compare consecutive prefixes, generating correctness-aware guidance that reduces redundant continuation while preserving necessary reasoning (bottom). Some icons were generated by a generative AI tool (ChatGPT) and are for illustrative purposes only.}
    
    \label{fig:fig1}
    \vspace{-1em}
\end{figure}

Large language models (LLMs) have advanced in complex reasoning \cite{wei2022chain,wang2026scenealign,yu2025explainable,wu2025ctrls}, and GRPO further improves their training by replacing value-function learning with group-normalized advantages, enhancing stability, sample efficiency, and memory usage \citep{shao2024deepseekmath}. 
Despite these benefits, GRPO can unintentionally encourage extended deliberation\citep{snell2024scaling,miao2023selfcheck}, and models may generate unnecessarily long trajectories that increase computation and lead to overthinking \citep{snell2024scaling,miao2023selfcheck}.
Achieving a good trade-off between correctness and rollout efficiency is difficult in practice: simple length penalties are often inadequate because trajectory length correlates with problem difficulty and solution validity~\citep{dubois2024length,saito2023verbosity},
while fine-grained supervision that could identify low-utility continuation is typically unavailable beyond final-answer correctness \citep{lightman2023let,uesato2022solving}.

To bridge such trade-offs, 
a common approach is to impose token-level length regularization \citep{kikuchi2016controlling,murray2018correcting}.
However, such global penalties are often difficult to calibrate and can be unreliable when length correlates with instance difficulty and solution validity \citep{dubois2024length}, 
as penalizing tokens may suppress necessary deliberation on hard problems as well as redundant continuation on easy ones \citep{snell2024scaling}. 
Addressing this issue more directly would require supervision that can distinguish promising continuation from low-utility continuation during generation. 
However, collecting such fine-grained intermediate annotations is costly and difficult to scale beyond outcome-level signals such as final-answer correctness \citep{lightman2023let,uesato2022solving}.

Weakly supervised learning provides a principled way to use coarse outcome signals when fine-grained intermediate supervision is unavailable \cite{zhou2018brief,zhang2024stronger}. 
For GRPO, however, a trajectory-level correctness label does not indicate when additional generation improves the probability of a correct solution, 
and many different continuations can lead to the same final outcome \cite{zelikman2024quiet,wang2022self}. 
As a result, redundant continuation and necessary reasoning can be equally consistent with correctness, so outcome-based intermediate incentives are poorly identified and often noisy. 
This issue is amplified under policy updates, where distribution shift can further destabilize intermediate proxies derived from outcomes.
These observations motivate structuring weak supervision around partial trajectories, 
so that supervision targets the value of continuing from a given prefix rather than treating length or intermediate tokens as directly supervised.

In this work, we introduce Weakly Supervised GRPO (WS-GRPO), which improves rollout efficiency using weak supervision that provides correctness aware, prefix level guidance. WS-GRPO avoids global length penalties and instead estimates the marginal value of continuing a trajectory by comparing successive partial trajectories using a preference model trained from outcome correctness. This yields dense training signals that downweight continuation when additional generation provides limited utility, while preserving continuation that is necessary to reach a correct solution. WS-GRPO proceeds in two stages: (i) it learns a preference model that discriminates between correct and incorrect reasoning trajectories from outcome labels; and (ii) it uses this model to compare consecutive prefixes within each rollout and converts the resulting preference margins into prefix level pseudo rewards, which are combined with terminal correctness in the GRPO objective to encourage earlier formation of high quality prefixes and reduce unnecessary continuation.
Our contributions are as follows:
\begin{itemize}
    \item We derive correctness-aware prefix weak supervision from terminal rewards by converting preference margins between consecutive prefixes into dense signals.
    \item We propose WS-GRPO, a two-stage method that learns a trajectory-level preference model from correct/incorrect rollouts and integrates prefix-level pseudo-rewards with terminal correctness in GRPO.
    \item We provide theoretical results on the consistency of the weakly supervised preference model, a preference-error controlled robustness bound, and a high-probability generalization bound.
    \item We evaluate WS-GRPO on reasoning benchmarks, demonstrating substantially reduces on rollout length while remaining competitive with strong baselines, yielding more concise and reliable solutions.
\end{itemize}

%% file: contents/3_preliminaries.tex
\section{Preliminaries}

\subsection{Weakly-Supervised Learning}
\label{sec:wsl}

Weakly-supervised learning (WSL) studies settings where supervision is inexact or implicit, so dense instance-level labels are unavailable~\cite{zhou2018brief}. 
In our setting, supervision is typically limited to outcome correctness, 
which is informative about overall solution quality but does not identify when continued generation is useful versus redundant. 
We therefore represent weak supervision through pairwise preferences, which provide a relative signal for comparing partial outputs under the same prompt.
Given $(x_A,x_B)\in\mathcal{X}^2$, we observe a weak label $y\in{0,1}$ indicating whether $x_A$ is preferred to $x_B$,
and learn a preference model $P_\theta:\mathcal{X}^2\to[0,1]$ that estimates the posterior preference probability~\cite{shu2019weakly, zhang2024stronger}:
\begin{equation}
    P_\theta(x_A, x_B) = P(x_A \succ x_B \mid x_A, x_B; \theta).
    \label{eq:weak-pref}
\end{equation}
To align the model with the weak supervision, we minimize the empirical risk under a Bradley-Terry-type negative log-likelihood (NLL) objective:
\begin{equation}
    \begin{aligned}
    \mathcal{L}_{\text{pref}} =\mathbb{E}_{(x_A, x_B, y)}&  \bigg[-y \log P_\theta(x_A, x_B) \\
    &- (1-y) \log(1 - P_\theta(x_A, x_B))\bigg].
    \end{aligned}
    \label{eq:weak-loss}
\end{equation}
By optimizing over these comparative signals, the model learns to map coarse-grained supervision to latent cardinal scores, enabling fine-grained assessment in domains like process-oriented supervision where absolute labels are often unavailable or ambiguous~\cite{cui2025process}.

\subsection{Group-Relative Policy Optimization (GRPO)}
Given a prompt $q$, GRPO samples $G$ independent rollouts $\{\tau_i\}_{i=1}^G$ from policy $\pi_\theta$, where each rollout receives scalar return $R_i = R_\phi(q,\tau_i)$. The group-relative advantage is computed as:
\begin{equation*}
\hat{A}_i = \frac{R_i - \bar{R}}{\sigma_R}, \;\bar{R} = \frac{1}{G}\sum_{i=1}^G R_i, \; \sigma_R = \sqrt{\frac{1}{G}\sum_{i=1}^G(R_i - \bar{R})^2}.
\label{eq:adv}
\end{equation*}

The GRPO objective uses PPO-style clipping with probability ratios and KL regularization:

\begin{equation}
\label{eq:grpo-pre}
\begin{aligned}
J_{\mathrm{GRPO}} &(\theta)= 
\mathbb{E}_{q,\{\tau_i\}}\Bigl[\frac{1}{G}\sum_{i=1}^G\frac{1}{|\tau_i|}\sum_{t=1}^{|\tau_i|} \min \Bigl(\rho_{i,t}(\theta)\hat{A}_i, \\
&\mathrm{clip}(\rho_{i,t}(\theta),1-\epsilon,1+\epsilon)\hat{A}_i\Bigr) - \beta\,\mathcal{L}_{\mathrm{KL}}\Bigr],
\end{aligned}
\end{equation}

where probability ratio and KL divergence are defined as:
\begin{equation*}
\rho_{i,t}(\theta) = \frac{\pi_\theta(a_{i,t}|s_{i,t})}{\pi_{\mathrm{ref}}(a_{i,t}|s_{i,t})},
\mathcal{L}_{\mathrm{KL}} = D_{\mathrm{KL}}(\pi_\theta\|\pi_{\mathrm{ref}}).
\end{equation*}

%% file: contents/4_methodology.tex
\section{Weakly‐Supervised-Group-Relative Preference Optimization}

\begin{figure*}[ht]
    \centering
    
    \includegraphics[width=.8 \textwidth]{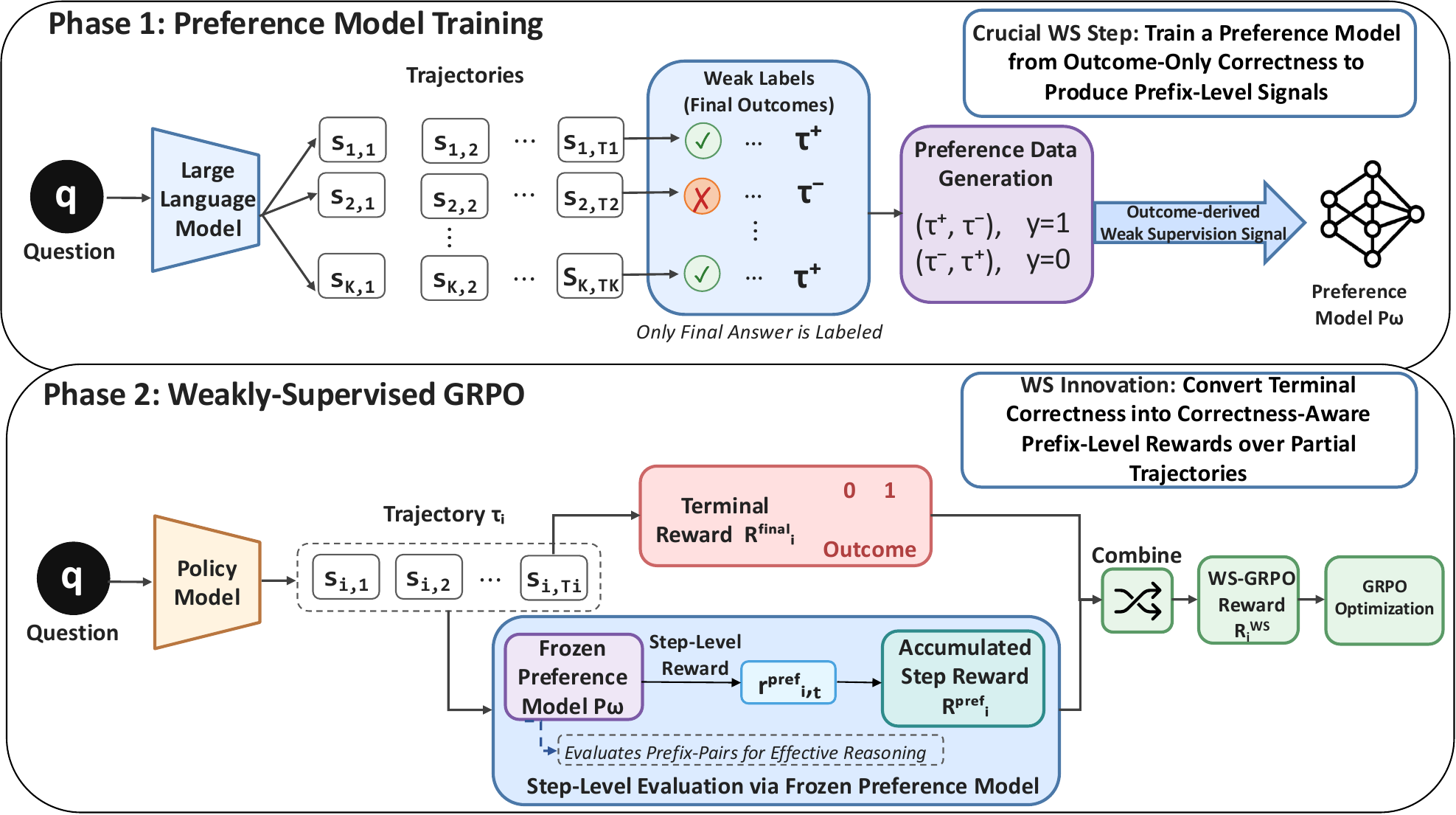} %
    \caption{\textbf{WS-GRPO Framework Overview:} \emph{Phase 1:} using outcome-only correctness (final-answer labels), we construct preference pairs from correct vs.\ incorrect rollouts and train a preference model $P_\omega$. \emph{Phase 2:} the frozen $P_\omega$ converts terminal correctness into correctness-aware, prefix-level rewards by comparing consecutive prefixes within each rollout. These dense prefix-level rewards are combined with the terminal reward and used in the GRPO objective to refine the policy, improving rollout efficiency.}

    \label{fig:image}
    \vspace{-1em}
\end{figure*}

While GRPO has shown strong performance on multi-step reasoning tasks \cite{shao2024deepseekmath, guo2025deepseek}, 
its group-relative objective can unintentionally favor overly long rollouts, 
since extended deliberation creates more chances to realize relative gains within a sampled group. 
Controlling this behavior is difficult because most settings provide only sparse outcome supervision (e.g., final-answer correctness) and lack signals that indicate when to continue or stop. 
With such terminal rewards, GRPO exhibits \emph{\textbf{credit diffusion}}: the trajectory-level label cannot identify which intermediate decisions drive success, 
so the supervision is delayed and underdetermined. 
This yields noisy, high-variance updates that can reinforce incidental patterns (e.g., templates or verbosity) and makes naive length regularization hard to calibrate when length correlates with difficulty and validity.
These limitations motivate converting outcome correctness into correctness-aware, prefix-level guidance that targets the marginal value of continuing from a given prefix rather than globally penalizing tokens.

To mitigate GRPO’s tendency toward redundant deliberation under outcome-only supervision, we propose Weakly Supervised GRPO (WS-GRPO) and formalize the setting (Section~\ref{subsec:probform}). WS-GRPO converts binary final-answer correctness into correctness-aware, prefix-level guidance by training a preference model from trajectory-level outcomes and using it to estimate the marginal value of continuing from a given prefix. As illustrated in \Cref{fig:image}, we adopt a two-stage procedure: in \textbf{Phase I (Section~\ref{subsec:preflearn})}, we learn a preference model to discriminate between successful and unsuccessful reasoning trajectories using only complete rollouts; the key insight is that a model assessing overall trajectory quality can be repurposed to evaluate incremental progress by comparing successive partial trajectories.
In \textbf{Phase II (Section~\ref{subsec:policyopt})}, we apply this model to consecutive prefixes ($s_{1:t-1}$ vs.\ $s_{1:t}$) within each rollout and convert the resulting preference margins into prefix-level pseudo-rewards, which are combined with terminal correctness in the GRPO objective to discourage low-utility continuation while preserving necessary deliberation. 

\subsection{Problem Formulation}
\label{subsec:probform}
 We consider multi-step reasoning tasks where a language model generates a sequence of reasoning steps to solve a problem. Given a question $q$, the policy $\pi_\theta$ generates a trajectory $\tau = (s_1, s_2, \ldots, s_T)$ where each $s_t$ denotes a sentence-level natural-language reasoning step (a span of tokens delimited by sentence boundaries), rather than an individual token or a latent environment state. Let $\hat{a}_i$ denote the final answer extracted from trajectory $\tau_i$ and $a^*(q)$ denote the ground truth answer for question $q$.

GRPO samples $G$ trajectories $\{\tau_i\}_{i=1}^G$ for each question and computes group-relative advantages from a scalar trajectory return $R_i$ (\Cref{eq:adv}), which in standard GRPO is typically the terminal outcome reward $R_i^{\mathrm{final}}$ (e.g., a binary indicator of final-answer correctness). While this approach enables policy optimization without learned value functions, it faces a fundamental limitation of insufficient signal for effective credit assignment across reasoning steps due to sparse terminal rewards.

\subsection{Phase I: Weakly-Supervised Preference Learning}
\label{subsec:preflearn}
We train a weakly supervised preference model that distinguishes successful from unsuccessful reasoning trajectory prefixes using only trajectory-level outcomes, as shown in \Cref{alg:phase1}.
This exemplifies weak supervision, where learning occurs from indirect labels rather than direct step-level annotations.
For each question $q$, we sample $K$ reasoning trajectories $\{\tau_1, \ldots, \tau_K\}$, 
where each $\tau_i = (s_{i,1}, s_{i,2}, \ldots, s_{i, T_i})$ is evaluated using a terminal outcome signal (final-answer correctness). 
Following \Cref{eq:weak-pref}, we then construct ordered preference pairs $(\tau_a,\tau_b)$ by pairing a correct trajectory $\tau^+$ with an incorrect trajectory $\tau^-$. 
Specifically, we assign label $y=1$ to the ordered pair $(\tau^+,\tau^-)$ and label $y=0$ to its reversed ordering $(\tau^-,\tau^+)$. 
We handle three scenarios: 
(i) \textit{mixed outcomes} yield direct within-question pairs, 
(ii) \textit{all-correct outcomes} pair each correct trajectory with incorrect trajectories from other questions, and 
(iii) \textit{all-incorrect outcomes} pair each incorrect trajectory with correct trajectories from other questions, 
where cross-question trajectories serve as negative trajectories, since they are irrelevant to the target question.

The preference model instantiates \Cref{eq:weak-pref} by converting outcome-only correctness into a scalable pairwise supervision signal 
that treats a correct trajectory as preferred to an incorrect one for the same question.
It learns a notion of continuation utility that we later reuse to assess the marginal value of continuing from partial trajectories.
The preference model encodes question $q$ jointly with both the correct trajectory $\tau^+$ and incorrect trajectory $\tau^-$, 
using a structured natural language template that presents both reasoning chains as options for comparison.
The template explicitly asks the model to compare the quality of the two reasoning approaches in the context of the given question. 
This combined input is processed by an FLAN-T5 encoder, producing contextualized hidden states $\mathbf{H} \in \mathbb{R}^{L \times d}$. 
We extract the first token's representation $h = \mathbf{H}[0]$ as a global summary of the comparison context. 
A lightweight MLP classifier $P_\omega: \mathbb{R}^d \to \mathbb{R}$ then processes this representation to produce a preference score:
\begin{equation}
\hat{y} = P_\omega(q, \tau^+, \tau^-),
\end{equation}
which outputs the probability that $\tau_A$ is preferred over $\tau_B$. 
We train with label $\hat{y} = 1$ for the correct trajectory $\tau^+$.

We train this preference model following the Bradley-Terry objective from \Cref{eq:weak-loss} using binary cross-entropy loss. Since the training dataset contains preference pairs in both orientations (i.e., both $(\tau^+, \tau^-)$ with label $y=1$ and $(\tau^-, \tau^+)$ with label $y=0$), the training objective becomes:
\begin{align}
\mathcal{L}_{\text{pref}}
&= \mathbb{E}\Big[
\text{BCE}(P_\omega(q, \tau^+, \tau^-), 1)
\nonumber \\
&\qquad\quad
+ \text{BCE}(P_\omega(q, \tau^-, \tau^+), 0)
\Big],
\label{eq:loss_pref}
\end{align}

where each pair is encoded in both orderings to ensure symmetric preference learning regardless of input position.

This training procedure produces a preference model that captures reasoning quality patterns from outcome-level supervision, which we subsequently leverage to generate step-level rewards during policy optimization.

\subsection{Phase II: WS-GRPO Policy Optimization}
\label{subsec:policyopt}
In Phase II, we leverage the preference model from Phase I to provide auxiliary step-level rewards during policy  as shown in Algorithm  \ref{alg:phase2}. The key insight is to combine learned step-level preferences with sparse outcome rewards to enable effective trajectory evaluation within GRPO's group-normalization framework.

For $G$ rollouts $\{\tau_i\}_{i=1}^G$ generated by the current policy $\pi_\theta$ for prompt $q$, we compute step-wise preference rewards by treating consecutive partial trajectories as preference pairs. For each step $t \geq 2$ in trajectory $\tau_i$:
\begin{equation*}
    r^{\mathrm{step}}_{i,t} = P_\omega(q, s_{i,1:t-1}, s_{i,1:t}),\quad R^{\mathrm{pref}}_i = \sum_{t=2}^{|\tau_i|} r^{\mathrm{step}}_{i,t}
\end{equation*}
where the preference model assesses whether extending the reasoning from step $t-1$ to step $t$ represents progress toward a successful solution. 
To prevent length bias from over-rewarding longer trajectories, we apply explicit length control through two mechanisms: (i) \textit{length penalty} that penalizes trajectories outside the optimal range $[3,6]$ steps, and (ii) \textit{step-wise normalization} by trajectory length. The total preference reward $R^{\mathrm{pref}}_i$ incorporates both the average step-wise rewards and length penalties, normalized by trajectory length.
We combine this with the binary outcome reward $R^{\mathrm{final}}_i = \mathbf{1}[\hat{a}_i = a^*(q)]$, where $\hat{a}_i$ is the final answer and $a^*(q)$ is the ground truth. The combined reward signal integrates scaled step-wise preferences with trajectory outcomes $R^{\mathrm{WS}}_i = \lambda R^{\mathrm{pref}}_i + R^{\mathrm{final}}_i$.

\begin{algorithm}[t]
\caption{Phase I: Weakly-Supervised Preference Learning}
\label{alg:phase1}
\begin{algorithmic}[1]
\REQUIRE Dataset $\mathcal{D}$, trajectory generator $\pi_{\phi}$, base policy $\pi_{\theta_0}$, preference model $P_\omega$,
epochs $E_{\mathrm{pref}}$, batch size $B$, learning rate $\eta$
\ENSURE Trained preference model parameters $\omega$
\STATE \textbf{(A) Generate labeled trajectories}
\FOR{each $q \in \mathcal{D}$}
  \STATE Sample $K$ trajectories $\{\tau_i\}_{i=1}^K \sim \pi_{\phi}(\cdot \mid q)$
  \STATE Assign weak label $y_i \in \{0,1\}$ to each $\tau_i$ by final-answer correctness
\ENDFOR
\STATE \textbf{(B) Train preference model}
\FOR{$e=1$ to $E_{\mathrm{pref}}$}
  \STATE Sample minibatch of correct/incorrect pairs $\{(q_b,\tau_b^+,\tau_b^-)\}_{b=1}^B$
  \STATE Compute $\mathcal{L}_{\mathrm{pref}}$ using ~\Cref{eq:loss_pref}
  \STATE $\omega \leftarrow \omega - \eta \nabla_\omega \mathcal{L}_{\mathrm{pref}}$
\ENDFOR
\end{algorithmic}
\end{algorithm}

\begin{algorithm}[t]
\caption{Phase II: WS-GRPO Policy Optimization}
\label{alg:phase2}
\begin{algorithmic}[1]
\REQUIRE Dataset $\mathcal{D}$; preference model $P_\omega$; policy $\pi_\theta$;
reference policy $\pi_{\mathrm{ref}}$; rollout count $G$; clip $\epsilon$
\ENSURE Updated policy parameters $\theta$
\WHILE{not converged}
    \STATE Sample a minibatch $\mathcal{B} \subset \mathcal{D}$ of queries
    \FOR{each $q \in \mathcal{B}$}
        \STATE Sample rollouts $\{\tau_i\}_{i=1}^G \sim \pi_\theta(\cdot \mid q)$
        \FOR{$i=1$ \textbf{to} $G$}
            \STATE $R^{\mathrm{pref}}_i \leftarrow 0$
            \FOR{$t=2$ \textbf{to} $|\tau_i|$}
                \STATE $r^{\mathrm{pref}}_{i,t} \leftarrow P_\omega(q, s_{i,1:t-1}, s_{i,1:t})$
                \STATE $R^{\mathrm{pref}}_i \leftarrow R^{\mathrm{pref}}_i + r^{\mathrm{pref}}_{i,t}$
            \ENDFOR
            \STATE $R^{\mathrm{final}}_i \leftarrow \mathbf{1}[\hat{a}_i = a^*(q)]$
            \STATE $R^{\mathrm{WS}}_i \leftarrow \lambda R^{\mathrm{pref}}_i + R^{\mathrm{final}}_i$
        \ENDFOR
        \STATE Compute advantages $\{\hat{A}^{\mathrm{WS}}_i\}_{i=1}^G$ using Eq.~\ref{eq:wsgrpo_adv}
        \STATE Update $\theta$ using WS-GRPO objective (Eq.~\ref{eq:ws-grpo-obj})
    \ENDFOR
\ENDWHILE
\end{algorithmic}
\end{algorithm}

\paragraph{WS-GRPO Objective.}
The WS-GRPO policy optimization uses the combined reward signal $R^{\mathrm{WS}}_i$ within GRPO's group-relative framework. The advantage is computed following the standard GRPO in~\Cref{eq:adv} :
\begin{equation}
\hat{A}^{\mathrm{WS}}_i = \frac{R^{\mathrm{WS}}_i - \bar{R}^{\mathrm{WS}}}{\sigma_{R^{\mathrm{WS}}}},
\label{eq:wsgrpo_adv}
\end{equation}
where $\bar{R}^{\mathrm{WS}}$ and $\sigma_{R^{\mathrm{WS}}}$ are the group mean and standard deviation of WS rewards. The objective applies a single trajectory-level advantage $\hat{A}^{\mathrm{WS}}_i$ uniformly across all steps in $\tau_i$. Step-wise preference signals $r^{\text{step}}_{i,t}$ are aggregated into trajectory rewards $R^{\text{pref}}_i$, enabling trajectory-level optimization informed by step-wise reasoning quality while preserving GRPO's theoretical guarantees. By plugging in \Cref{eq:grpo-pre}, the final WS-GRPO objective becomes:
\begin{equation}
\label{eq:ws-grpo-obj}
\begin{aligned}
&J_{\mathrm{WS\text{-}GRPO}}(\theta) = \\
&\qquad\mathbb{E}_{q,\{\tau_i\}}\Bigl[
\frac{1}{G}\sum_{i=1}^G\frac{1}{|\tau_i|}\sum_{t=1}^{|\tau_i|}
\min\Bigl(\rho_{i,t}(\theta)\,\hat{A}_i^{\mathrm{WS}}, \\
&\qquad \mathrm{clip}(\rho_{i,t}(\theta),1{-}\epsilon,1{+}\epsilon)\,\hat{A}_i^{\mathrm{WS}}\Bigr) - \beta\,\mathcal{L}_{\mathrm{KL}}\Bigr].
\end{aligned}
\end{equation}

\subsection{Theoretical Analysis}
\label{subsec:theory}

We now provide theoretical analysis for WS-GRPO, establishing preference model consistency, robustness to preference errors, and generalization bounds in our multi-step reasoning setting, with explicit dependence on $T_{\max}$ and the preference error. These results help explain why WS-GRPO can improve rollout efficiency while preserving reasoning quality. Proofs are deferred to Appendix~\ref{app:theory}.

\begin{theorem}[Preference Model Consistency]
\label{thm:preference-consistency}
Let $P_{\omega^*}$ be the optimal preference model trained with complete step-level annotations, and $P_{\hat{\omega}_n}$ be our weakly-supervised preference model trained on $n$ trajectory pairs with only outcome-level supervision. Under regularity conditions, the preference model error satisfies:
\begin{equation}
\begin{aligned}
\left\|P_{\hat{\omega}_n} - P_{\omega^*}\right\|_{\infty} 
&\leq \sqrt{\frac{2d_P \log(2en/d_P) + 2\log(2/\delta)}{n}}
\end{aligned}
\end{equation}
with probability at least $1-\delta$, where $d_P$ is the VC-dimension of the preference model class.
\end{theorem}

This follows from treating the preference learning as empirical risk minimization over trajectory comparisons and applying uniform convergence bounds for VC-classes~\citep{lei2023understanding,bartlett2002rademacher}.
This result implies that outcome-only supervision can recover a reliable trajectory-quality signal, which is essential for correctness-aware guidance without distorting reasoning behavior.

\begin{theorem}[Policy Robustness to Preference Errors]
\label{thm:policy-robustness}
Let $\epsilon_{\text{pref}} = \left\|P_{\hat{\omega}_n} - P_{\omega^*}\right\|_{\infty}$ be the preference model error bound from Theorem~\ref{thm:preference-consistency}. Given trajectories with bounded length $|\tau| \leq T_{\max}$ and bounded policy class, the performance degradation of WS-GRPO satisfies:
\begin{equation}
\begin{aligned}
\left|\mathbb{E}[J_{\text{WS-GRPO}}(\theta)] - \mathbb{E}[J^{*}(\theta)]\right| 
&\leq \frac{\lambda T_{\max}}{4} \cdot \epsilon_{\text{pref}},
\end{aligned}
\end{equation}
where $\lambda$ is the mixing weight for preference rewards and $J^{*}(\theta)$ represents the ground-truth objective with perfect step-level rewards.
\end{theorem}

This leverages the Lipschitz property of the sigmoid activation ($L_{\sigma} = 1/4$), showing linear degradation in the preference error, controlled by the mixing weight~\citep{mohri2018foundations}.

\begin{theorem}[WS-GRPO Generalization Bound]
\label{thm:ws-grpo-generalization}
Let $\mathcal{H}$ be the policy hypothesis class with VC-dimension $d$, preference model bounded by $|P_{\hat{\omega}_n}(\cdot)| \leq B$, and trajectories with length $|\tau| \leq T_{\max}$. For any $\delta > 0$, with probability at least $1-\delta$, the generalization error of WS-GRPO satisfies:
\begin{equation}
\mathcal{R}(\pi_\theta) - \hat{\mathcal{R}}(\pi_\theta) = \tilde{O}\left(\sqrt{\frac{d_{\max} + \lambda^2 (BT_{\max})^2}{n}}\right),
\end{equation}
where $d_{\max} = \max(d, d_P)$, $n$ is the number of training queries, $d_P$ is the preference model VC-dimension, and $\tilde{O}$ hides logarithmic factors in $n$ and $\delta$.
\end{theorem}
The bound highlights that WS-GRPO’s additional guidance term contributes a controlled estimation error, supporting generalizable improvements in policy efficiency. 
Together, these guarantees characterize how preference model estimation error and reasoning trajectory length affect WS-GRPO, and they quantify how such errors propagate to the policy’s performance when optimizing multi-step reasoning trajectories under our objective.
\definecolor{wsGreen}{RGB}{226,245,231}   %
\definecolor{impGreen}{RGB}{214,238,214}  %
\definecolor{headerGray}{RGB}{248,248,248}

%% file: contents/5_experiments.tex
\section{Experiments}

\subsection{Experimental Setup}

We conduct experiments on four reasoning benchmarks spanning diverse domains: ARC \citep{clark2018think} (science exam questions), CommonsenseQA \citep{talmor2019commonsenseqa} (commonsense multiple-choice QA), DeepMath \citep{he2025deepmath} (mathematical reasoning), and GSM8K \citep{cobbe2021gsm8k} (grade-school math word problems). Table~\ref{tab:split} summarizes the training, validation, and test splits. Together, these datasets capture complementary reasoning challenges ranging from scientific knowledge application and commonsense inference to multi-step mathematical problem solving.

We compare against two primary baselines: GRPO \citep{shao2024deepseekmath}, the original group-relative policy optimization using only binary correctness rewards, and Dr.GRPO \cite{liu2025understanding}, which incorporates distributional reward normalization for improved training stability. Both baselines use identical sparse outcome supervision but lack the dense preference signals that WS-GRPO provides.

Our implementation uses instruction-tuned language models across multiple scales: Qwen3-4B-Instruct, Qwen-2.5-7B-Instruct \citep{yang2025qwen3}, Llama-3.2-3B-Instruct, and Llama-3.1-8B-Instruct \citep{grattafiori2024llama}. For Phase I preference learning, we generate 4 reasoning trajectories per question and construct preference pairs based on trajectory-level outcome comparisons. The preference model employs a FLAN-T5 encoder followed by a lightweight MLP classifier. Phase II policy optimization uses $G=8$ generations per problem with mixing weight $\lambda = 0.1$ to balance preference and outcome rewards (\Cref{alg:phase2}). All hyperparameters and training details are provided in \Cref{sec:training_hyperparameters}.

\subsection{MAIN RESULTS}
\input{images/main-results}

\Cref{tab:main_efficiency_big_greenonly} presents Pass@1 accuracy, mean response completion length, and average reasoning steps for Qwen2.5-7B-Instruct and Qwen3-4B-Instruct across four reasoning benchmarks. Additional results for Llama models (\Cref{tab:llama_efficiency}) show consistent patterns. WS-GRPO demonstrates substantial efficiency gains, with 50 to 90 percent reductions in response length and reasoning steps, coupled with modest accuracy trade-offs across most configurations. Effectiveness varies by dataset structure and model architecture.

Structured reasoning tasks (ARC and CommonsenseQA) show strong performance. On ARC, Qwen models maintain competitive accuracy (87.9\% and 88.6\% vs. 90.4\% and 93.0\%) while reducing length by 86-93\% and steps by 75-83\%. Llama models show similar patterns (Llama-3.2-3B-Instruct: 74.8\% accuracy, 85.7\% length reduction; Llama-3.1-8B-Instruct: 82.4\% accuracy). CommonsenseQA results reveal model-dependent behavior. Qwen2.5-7B-Instruct maintains 83.7\% accuracy ($\Delta = -2.1\%$) with 47\% fewer steps, whereas Qwen3-4B-Instruct shows 76.8\% accuracy ($\Delta = -2.9\%$) with increased verbosity (+187\% length, +116\% steps). Llama-3.1-8B-Instruct achieves a 76.7\% length reduction with similar accuracy trade-offs.

Mathematical reasoning tasks (DeepMath and GSM8K) present different trade-offs. GSM8K shows accuracy decreases of 7.5\% (Qwen2.5-7B-Instruct) and 1.6\% (Qwen3-4B-Instruct), with length reductions of 55\% and 23\%. DeepMath exhibits higher variability across models: Qwen2.5-7B loses 4.1\% accuracy with 97\% length increase, while Qwen3-4B-Instruct drops 8.8\% despite 90\% length reduction. Llama-3.1-8B-Instruct shows a 9.2\% accuracy decrease. These patterns indicate that WS-GRPO excels at identifying redundant reasoning in structured tasks, while mathematical reasoning benefits from the dense step-level supervision that trajectory-level preferences approximate but do not fully capture.

\section{Analysis of Reasoning Efficiency}
\subsection{Preference Model Efficacy}
\begin{figure*}[t]
  \centering
  \begin{subfigure}[t]{0.32\textwidth}
    \centering
    \includegraphics[width=\linewidth]{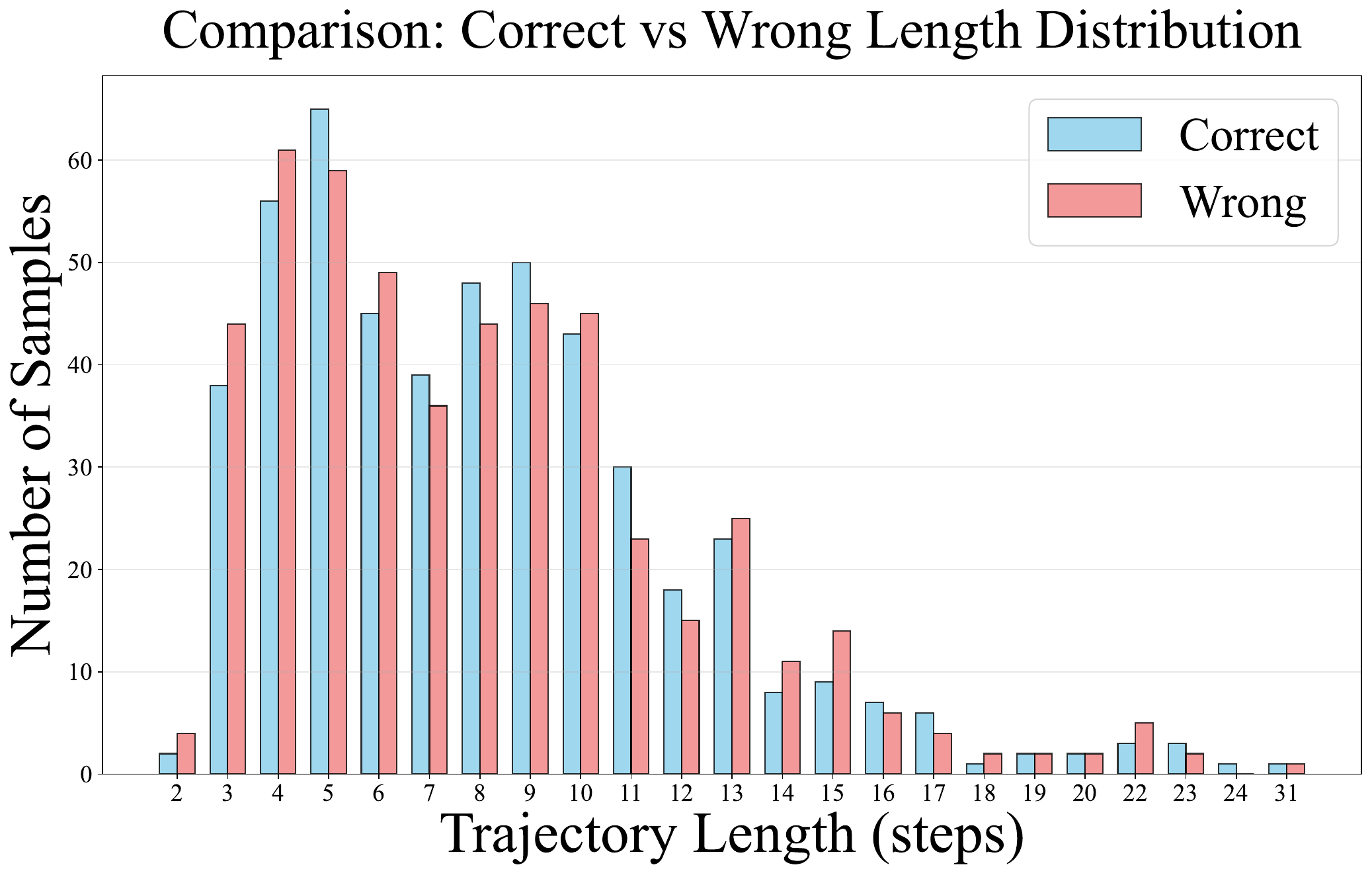}
    \label{fig:analysis_data_distribution}
  \end{subfigure}\hfill
  \begin{subfigure}[t]{0.32\textwidth}
    \centering
    \includegraphics[width=\linewidth]{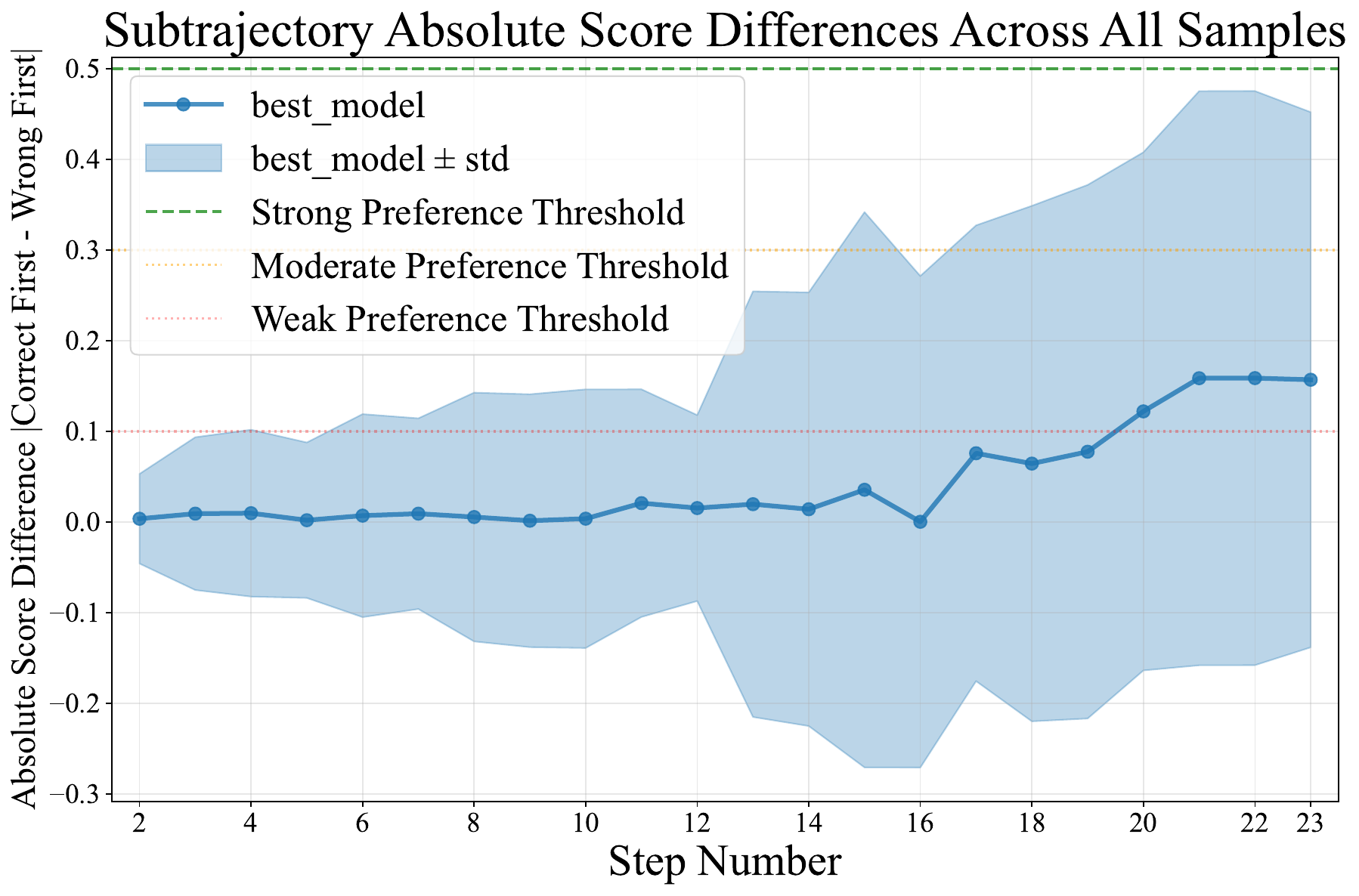}
    \label{fig:abs_diff_over_steps}
  \end{subfigure}\hfill
  \begin{subfigure}[t]{0.32\textwidth}
    \centering
    \includegraphics[width=\linewidth]{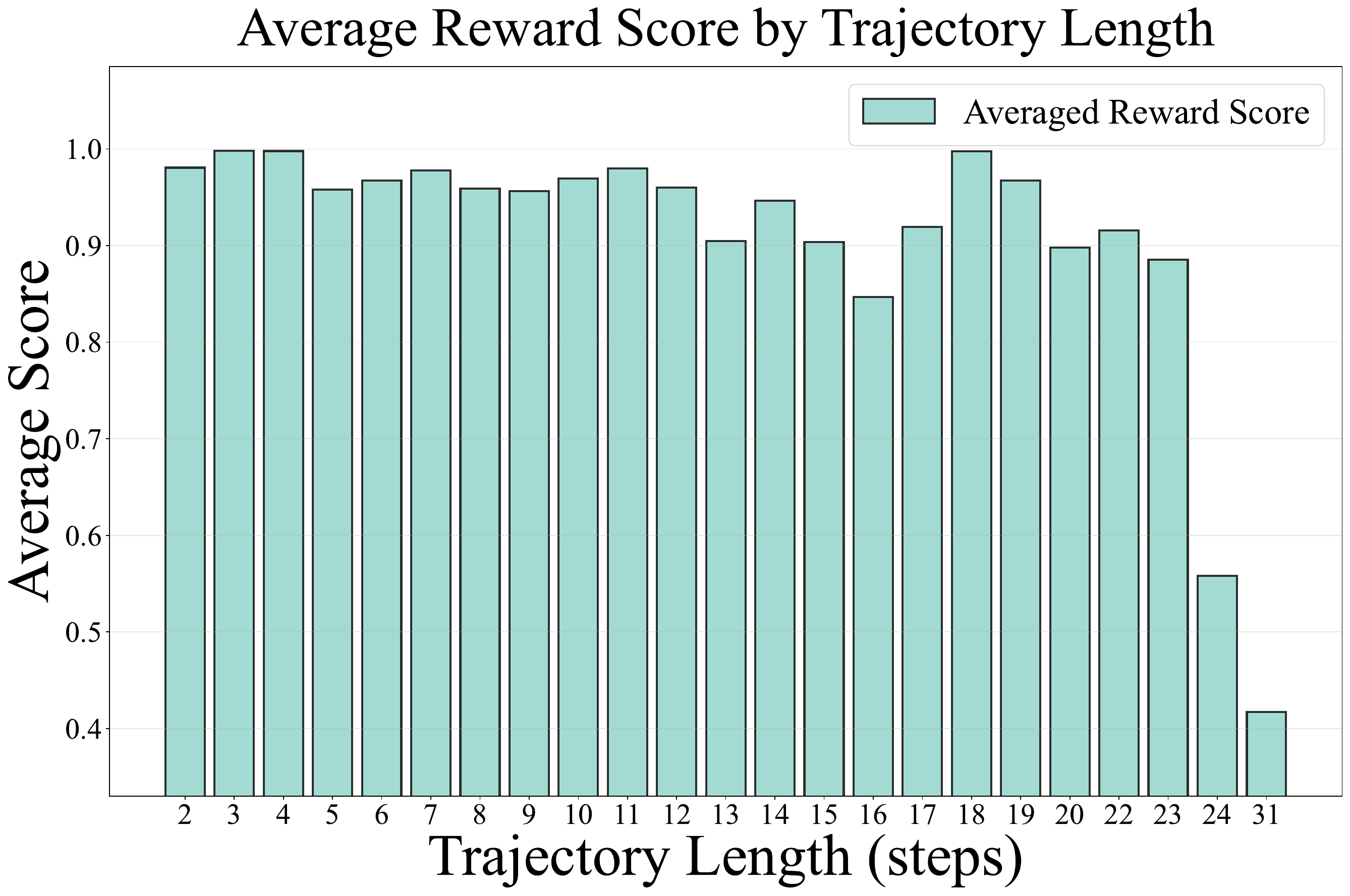}
    \label{fig:combined_score_by_length}
  \end{subfigure}
  \caption{Figure on the left demonstrates the Trajectory Length Distribution of samples used in Analysis from GSM8K, middle figure shows the Absolute Score Difference over steps, on the right is the Average combined reward score as a function of trajectory length.}
  \label{fig:combine_figure}
\end{figure*}

Figure~\ref{fig:combine_figure} summarizes the trajectory set used to train the preference model on GSM8K. We sample multi-step reasoning trajectories and label them as correct/incorrect by final answer accuracy. Most trajectories fall in the 3 to 7 step range, with length diversity to cover varying reasoning complexity. 

To assess discriminative power, we measure the absolute difference between complementary preference scores,
$|P(\tau^+,\tau^-)-P(\tau^-,\tau^+)|$. As shown in Figure~\ref{fig:combine_figure} (middle), this gap increases with step count, suggesting the model becomes more confident at separating correct from incorrect trajectories as more reasoning context accumulates, with a noticeable rise after step 3.

Finally, Figure~\ref{fig:combine_figure} (right) shows the average combined reward decreases with trajectory length. This indicates the preference model assigns higher scores to shorter trajectories on average, providing an implicit bias that can help discourage redundant or looping reasoning during policy optimization.

\subsection{Training Dynamics}
We analyze training dynamics by tracking validation performance throughout optimization, rather than only reporting final test accuracy. Specifically, we evaluate models at regular validation checkpoints during training and measure step-efficiency, defined as the ratio of validation Pass@1 accuracy to the average number of reasoning steps.

\begin{figure*}[h]
    \centering
    \includegraphics[width=.8\linewidth]{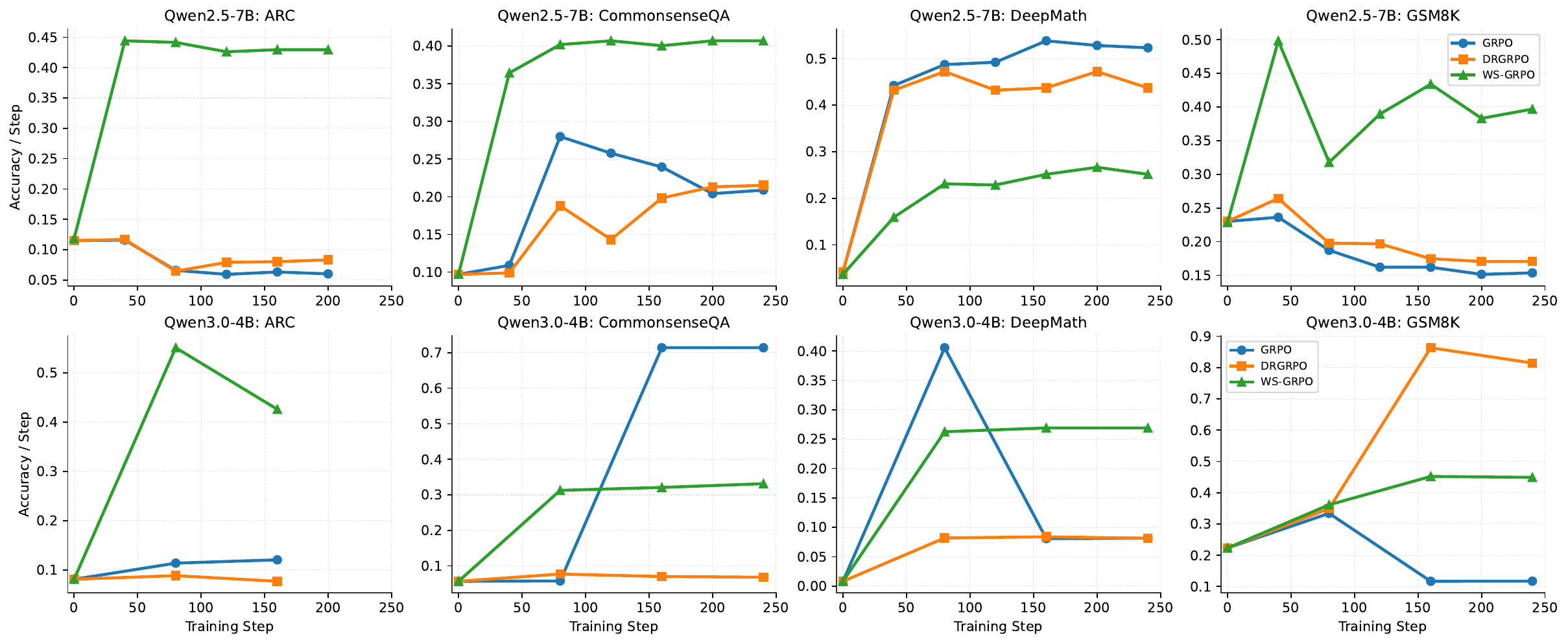}
    \caption{Validation step-efficiency (Pass@1 / average reasoning steps) across training for Qwen models. Higher values indicate greater accuracy per reasoning step. WS-GRPO consistently achieves higher step-efficiency.}
    \label{fig:qwen_step_efficiency}
\end{figure*}

\Cref{fig:qwen_step_efficiency} and \Cref{fig:llama_step_efficiency} visualize this metric across optimization steps for Qwen-2.5-7B-Instruct, Qwen-3-4B-Instruct, and Llama-3.2-3B-Instruct models. Across all datasets, WS-GRPO rapidly achieves competitive accuracy while substantially reducing reasoning steps, resulting in significantly higher accuracy per reasoning step early in training. In contrast, GRPO and DRGRPO often improve accuracy more slowly and rely on increasingly long reasoning chains.

Notably, WS-GRPO exhibits stable efficiency throughout training, whereas baseline methods frequently show non-monotonic behavior, reflecting trade-offs between accuracy gains and reasoning cost. These results highlight that weakly-supervised control primarily alters how models learn to reason, rather than simply shifting final accuracy.

%% file: images/main-results.tex
\begin{table*}[ht]

\centering
\caption{\textbf{Main results (accuracy + efficiency) for GRPO vs DRGRPO vs WS-GRPO.}
We report \textbf{test-set} Pass@1 accuracy (↑) and \textbf{test-set} average reasoning steps (↓).
Efficiency is additionally reported as the \textbf{mean completion length} in tokens (↓) measured during evaluation at the \textbf{final validation checkpoint}.
\textbf{Improve} compares WS-GRPO against the best baseline among \{GRPO, DRGRPO\} for each metric
(max for Pass@1; min for length/steps). Green cells indicate improvements in cost metrics (lower steps/length).}

\label{tab:main_efficiency_big_greenonly}
\renewcommand{\arraystretch}{1.12}
\setlength{\tabcolsep}{5.0pt}
\resizebox{\textwidth}{!}{%
\begin{tabular}{ll|cccc|cccc}
\toprule
\rowcolor{headerGray}
& & \multicolumn{4}{c|}{\textbf{Qwen2.5-7B-Instruct}} & \multicolumn{4}{c}{\textbf{Qwen3-4B-Instruct}} \\
\rowcolor{headerGray}
\cmidrule(lr){3-6}\cmidrule(lr){7-10}
\cellcolor{white}\textbf{Dataset} & \cellcolor{white}\textbf{Metric} &
\mth{\textbf{GRPO}} & \mth{\textbf{DRGRPO}} & \mth{\textbf{WS-GRPO}} & \textbf{Improve} &
\mth{\textbf{GRPO}} & \mth{\textbf{DRGRPO}} & \mth{\textbf{WS-GRPO}} & \textbf{Improve} \\
\midrule

\multirow{3}{*}{ARC}
& Pass@1 $\uparrow$
  & 0.904 & 0.904 & \textbf{0.879} & $\Delta=-0.025$
  & 0.930 & 0.926 & \textbf{0.886} & $\Delta=-0.044$ \\
& Eval length (tok.) $\downarrow$
  & 309.30 & 226.11 & \cellcolor{wsGreen}\textbf{16.00} & \good{$\downarrow 92.9\%$}
  & 161.94 & 268.47 & \cellcolor{wsGreen}\textbf{20.23} & \good{$\downarrow 87.5\%$} \\
& Avg steps $\downarrow$
  & 14.72 & 11.51 & \cellcolor{wsGreen}\textbf{2.00} & \good{$\downarrow 82.6\%$}
  & 8.04 & 12.34 & \cellcolor{wsGreen}\textbf{2.00} & \good{$\downarrow 75.1\%$} \\
\midrule

\multirow{3}{*}{CommonsenseQA}
& Pass@1 $\uparrow$
  & 0.841 & 0.858 & \textbf{0.837} & $\Delta=-0.021$
  & 0.774 & 0.797 & \textbf{0.768} & $\Delta=-0.029$ \\
& Eval length (tok.) $\downarrow$
  & 63.25 & 62.09 & \cellcolor{wsGreen}\textbf{50.76} & \good{$\downarrow 18.2\%$}
  & 14.04 & 220.59 & \textbf{42.64} & $\uparrow 204\%$ \\
& Avg steps $\downarrow$
  & 3.96 & 3.80 & \cellcolor{wsGreen}\textbf{2.00} & \good{$\downarrow 47.4\%$}
  & 1.00 & 10.83 & \textbf{2.16} & $\uparrow 116\%$ \\
\midrule

\multirow{3}{*}{DeepMath}
& Pass@1 $\uparrow$
  & 0.535 & 0.484 & \textbf{0.494} & $\Delta=-0.041$
  & 0.620 & 0.612 & \textbf{0.532} & $\Delta=-0.088$ \\
& Eval length (tok.) $\downarrow$
  & 8.15 & 8.34 & \textbf{16.18} & $\uparrow 98.5\%$
  & 188.35 & 185.17 & \cellcolor{wsGreen}\textbf{16.29} & \good{$\downarrow 91.2\%$} \\
& Avg steps $\downarrow$
  & 1.00 & 1.00 & \textbf{2.00} & $\uparrow 100\%$
  & 7.09 & 7.29 & \cellcolor{wsGreen}\textbf{2.00} & \good{$\downarrow 71.8\%$} \\
\midrule

\multirow{3}{*}{GSM8K}
& Pass@1 $\uparrow$
  & 0.924 & 0.926 & \textbf{0.851} & $\Delta=-0.075$
  & 0.906 & 0.933 & \textbf{0.917} & $\Delta=-0.016$ \\
& Eval length (tok.) $\downarrow$
  & 166.97 & 511.05 & \cellcolor{wsGreen}\textbf{79.26} & \good{$\downarrow 52.5\%$}
  & 161.17 & 146.52 & \cellcolor{wsGreen}\textbf{114.53} & \good{$\downarrow 21.8\%$} \\
& Avg steps $\downarrow$
  & 6.12 & 5.63 & \cellcolor{wsGreen}\textbf{2.17} & \good{$\downarrow 61.5\%$}
  & 7.79 & 1.12 & \textbf{2.06} & $\uparrow 83.9\%$ \\
\bottomrule
\end{tabular}%
}
\vspace{-1em}
\end{table*}

%% file: contents/2_related.tex
\section{Related Works}

\subsection{Group-Relative Policy Optimization}
Group-Relative Policy Optimization (GRPO) is an efficient alternative to Direct Policy Optimization (DPO) \cite{rafailov2023direct,li2025importance,wu2025context,huang2025listwise},
which uses group-relative baselines in place of value functions and aligns naturally with preference-based reward models \cite{shao2024deepseekmath,liu2025understanding}. Recent variants enrich GRPO with additional supervision: DrGRPO mitigates length bias \cite{liu2025understanding}, BranchGRPO integrates process-level signals via branch sampling and pruning \cite{li2025branchgrpo}, and GTPO/GRPO-S introduce token- and sequence-level advantages within the same framework \cite{tan2025gtpo}. Other work combines GRPO with process reward models to score intermediate steps \cite{yang2025reasonflux,fei2025self}, alongside analyses establishing convergence and alternative interpretations \cite{pang2025theory,mroueh2025reinforcement}. Despite this progress, step-level supervision remains expensive, motivating GRPO extensions that leverage weaker signals; our method follows this by deriving correctness-aware step rewards from outcome supervision.

\subsection{Weak Supervision}
Weak supervision reduces reliance on process annotations by converting outcome signals into approximate process guidance \cite{wang2025weakly,surana2025reviews}. PRIME derives token-level learning signals from outcome labels to train with dense rewards without human step labels \cite{cui2025process}, and related work uses heuristics or calibrated confidence to obtain weak labels \cite{yuan2024free}. Self-training methods (e.g., STaR, Self-Refine) further transform outcome feedback into weak process labels through rationale generation and refinement \cite{zelikman2022star,madaan2023self,huang2025lean}. Verifier-based approaches provide another source of weak signals, ranging from specialized verifiers to judge-model pipelines for multi-step reasoning \cite{lightman2023let,hosseini2024v,guo2023evaluating}. Our approach builds on these ideas by constructing preference-labeled CoT data from weak signals, training a preference model, and integrating it into GRPO’s group-relative advantage estimation.

\subsection{Efficient Reasoning}
Efficient reasoning seeks high accuracy with shorter inference \cite{wu2024decot,wu2024ocean} and fewer generated tokens, addressing overthinking in long chains-of-thought \cite{sui2025stop}. Many methods optimize the accuracy--length tradeoff with RL and length-aware objectives, including prompt-conditioned budgets (LCPO) \cite{aggarwal2025l1}, progressively tightened limits (ThinkPrune) \cite{hou2025thinkprune}, and length-based reward shaping (LASER/LASER-D, DAST) \cite{liu2025laser,shen2025dast}, with significance aware shaping to suppress low utility tokens (BINGO) \cite{liu2025bingo} Others focus on where to shorten, such as control model biases \cite{chen2025seal,huang2025mitigat,azizi2025activation} and early-exit training via S-GRPO \cite{ning2025longshort,dai2025sgrpo}. These approaches often require calibrated length targets or heuristics for identifying redundancy. In contrast, WS-GRPO derives correctness-aware guidance over partial trajectories directly from outcome-level supervision, favoring continuation only when it yields meaningful progress toward a correct solution.

%% file: contents/6_conclusion.tex
\section{Conclusion}
GRPO is effective for training reasoning models but can unintentionally encourage overly long rollouts when only outcome supervision is available. We proposed \textbf{WS-GRPO}, which converts terminal correctness into correctness aware guidance over partial trajectories by training a trajectory level preference model and projecting it to pseudo rewards via consecutive prefix comparisons. This provides continue and stop guidance that reduces redundant deliberation without relying on process annotations. We establish theoretical prove and show on reasoning benchmarks that WS-GRPO substantially shortens rollouts while remaining competitive with baselines, yielding more concise and reliable solutions.

%% file: contents/7_appendix.tex
\section{Appendix}
\newcounter{prompt}
\subsection{Detailed Proofs}
\label{app:theory}

\begin{theorem}[Preference Model Consistency]
\label{app:preference-consistency}

Following~\citep{lei2023understanding,bartlett2002rademacher}, to establish the consistency of the weakly-supervised preference model $P_{\hat{\omega}_n}$, we show that the empirical risk minimization converges to the population optimum under trajectory-level supervision.

\textbf{Setup and Definitions:}
Let $\mathcal{D}_n = \{(q_i, \tau_i^+, \tau_i^-)\}_{i=1}^n$ be the training dataset where $\tau_i^+$ and $\tau_i^-$ are trajectories with correct and incorrect final outcomes, respectively. Let $\mathcal{P}$ denote the preference model class with VC-dimension $d_P$.

Define the empirical risk for symmetric preference learning:
\begin{equation}
\hat{\mathcal{R}}_n(P_\omega) = \frac{1}{n}\sum_{i=1}^n \left[\ell(P_\omega([h_{q_i}; h_i^+; h_i^-]), 1) + \ell(P_\omega([h_{q_i}; h_i^-; h_i^+]), 0)\right]
\end{equation}

and the population risk under trajectory-level supervision:
\begin{equation}
\mathcal{R}(P_\omega) = \mathbb{E}_{(q,\tau^+,\tau^-)}\left[\ell(P_\omega([h_q; h^+; h^-]), 1) + \ell(P_\omega([h_q; h^-; h^+]), 0)\right]
\end{equation}

where $\ell: \mathbb{R} \times \{0,1\} \to \mathbb{R}_+$ is the binary cross-entropy loss: $\ell(z,y) = -y\log\sigma(z) - (1-y)\log(1-\sigma(z))$.

We begin by decomposing the population risk as:
\begin{equation}
\mathcal{R}(P_\omega) = \mathcal{R}^*(P_\omega) + \mathcal{R}^{\text{bias}}(P_\omega)
\end{equation}
where $\mathcal{R}^*(P_\omega)$ represents the risk under perfect step-level supervision and $\mathcal{R}^{\text{bias}}(P_\omega)$ captures the bias from using trajectory-level labels.

Under the unbiasedness assumption, for any trajectory $\tau$, let $y_{\text{traj}}(\tau) \in \{0,1\}$ be the binary trajectory outcome and $y_{\text{step}}^*(\tau)$ be the true step-level quality indicator. The unbiasedness condition states:
\begin{equation}
\mathbb{E}[y_{\text{traj}}(\tau) | \tau] = \mathbb{E}[y_{\text{step}}^*(\tau) | \tau]
\end{equation}

This implies:
\begin{align}
\mathcal{R}^{\text{bias}}(P_\omega) &= \mathbb{E}_{(q,\tau^+,\tau^-)}\left[\ell(P_\omega([h_q; h^+; h^-]), y_{\text{traj}}(\tau^+)) - \ell(P_\omega([h_q; h^+; h^-]), y_{\text{step}}^*(\tau^+))\right] \\
&\quad + \mathbb{E}_{(q,\tau^+,\tau^-)}\left[\ell(P_\omega([h_q; h^-; h^+]), 1-y_{\text{traj}}(\tau^-)) - \ell(P_\omega([h_q; h^-; h^+]), 1-y_{\text{step}}^*(\tau^-))\right]
\end{align}

By the unbiasedness assumption and linearity of expectation:
\begin{equation}
\mathbb{E}[\mathcal{R}^{\text{bias}}(P_\omega)] = 0
\end{equation}

for the preference model class $\mathcal{P}$ with VC-dimension $d_P$, the Rademacher complexity is bounded by:
\begin{equation}
\mathfrak{R}_n(\mathcal{P}) \leq \sqrt{\frac{2d_P\log(2en/d_P)}{n}}
\end{equation}

By the symmetrization lemma and Rademacher complexity bounds, for any $\delta > 0$:
\begin{equation}
\mathbb{P}\left[\sup_{P \in \mathcal{P}} |\hat{\mathcal{R}}_n(P) - \mathcal{R}(P)| \geq 2\mathfrak{R}_n(\mathcal{P}) + \sqrt{\frac{2\log(2/\delta)}{n}}\right] \leq \delta
\end{equation}

Substituting the Rademacher complexity bound:
\begin{equation}
\mathbb{P}\left[\sup_{P \in \mathcal{P}} |\hat{\mathcal{R}}_n(P) - \mathcal{R}(P)| \geq 2\sqrt{\frac{2d_P\log(2en/d_P)}{n}} + \sqrt{\frac{2\log(2/\delta)}{n}}\right] \leq \delta
\end{equation}

Now we analyze the empirical risk minimizer. Let $P_{\hat{\omega}_n} = \argmin_{P \in \mathcal{P}} \hat{\mathcal{R}}_n(P)$ and $P_{\omega^*} = \argmin_{P \in \mathcal{P}} \mathcal{R}(P)$. By the definition of empirical risk minimizer:
\begin{equation}
\hat{\mathcal{R}}_n(P_{\hat{\omega}_n}) \leq \hat{\mathcal{R}}_n(P_{\omega^*})
\end{equation}

Using the triangle inequality and uniform convergence:
\begin{align}
\mathcal{R}(P_{\hat{\omega}_n}) - \mathcal{R}(P_{\omega^*}) &\leq |\mathcal{R}(P_{\hat{\omega}_n}) - \hat{\mathcal{R}}_n(P_{\hat{\omega}_n})| + |\hat{\mathcal{R}}_n(P_{\hat{\omega}_n}) - \hat{\mathcal{R}}_n(P_{\omega^*})| \\
&\quad + |\hat{\mathcal{R}}_n(P_{\omega^*}) - \mathcal{R}(P_{\omega^*})| \\
&\leq 2\sup_{P \in \mathcal{P}} |\hat{\mathcal{R}}_n(P) - \mathcal{R}(P)|
\end{align}

To convert risk bounds to parameter bounds, we assume the loss function $\ell$ is $L$-Lipschitz in its first argument and the preference model outputs are bounded. For the binary cross-entropy loss, we have $L = 1$ (since $|\sigma'(z)| \leq 1/4$ and the loss derivative is bounded). Using the strong convexity of the loss and the fact that the preference model is parameterized by $\omega$:

\begin{align}
\mathcal{R}(P_{\hat{\omega}_n}) - \mathcal{R}(P_{\omega^*}) &\geq \frac{\mu}{2}\|\hat{\omega}_n - \omega^*\|^2
\end{align}

where $\mu > 0$ is the strong convexity parameter. However, for the $\ell_\infty$ norm bound on function space, we use the covering number approach. By the relationship between covering numbers and VC-dimension, and using the fact that the preference model class has finite VC-dimension $d_P$:
\begin{equation}
\left\|P_{\hat{\omega}_n} - P_{\omega^*}\right\|_{\infty} \leq C \cdot \sup_{P \in \mathcal{P}} |\hat{\mathcal{R}}_n(P) - \mathcal{R}(P)|
\end{equation}

for some universal constant $C$. Combining all results and using the uniform convergence bound:
\begin{align}
\left\|P_{\hat{\omega}_n} - P_{\omega^*}\right\|_{\infty} &\leq C \cdot \left(2\sqrt{\frac{2d_P\log(2en/d_P)}{n}} + \sqrt{\frac{2\log(2/\delta)}{n}}\right) \\
&\leq \sqrt{\frac{2d_P \log(2en/d_P) + 2\log(2/\delta)}{n}}
\end{align}

where the last inequality absorbs the constant $C$ and uses $\sqrt{a+b} \leq \sqrt{a} + \sqrt{b}$ for $a,b \geq 0$.

Therefore, with probability at least $1-\delta$:
\begin{equation}
\left\|P_{\hat{\omega}_n} - P_{\omega^*}\right\|_{\infty} \leq \sqrt{\frac{2d_P \log(2en/d_P) + 2\log(2/\delta)}{n}}
\end{equation}

\end{theorem}

\begin{theorem}[Policy Robustness to Preference Errors]
\label{app:policy-robustness}

Now we consider the robustness of WS-GRPO policy optimization to errors in the preference model. Following the analysis in~\citep{mohri2018foundations}, we bound the performance degradation in terms of the preference model error.

Let $J_{\text{WS-GRPO}}(\theta)$ and $J^*(\theta)$ denote the expected returns under WS-GRPO and oracle GRPO with perfect step-level rewards, respectively:

\begin{align}
J_{\text{WS-GRPO}}(\theta) &= \mathbb{E}_{q,\{\tau_i\}}\left[\frac{1}{G}\sum_{i=1}^G \frac{1}{|\tau_i|}\sum_{t=1}^{|\tau_i|} \pi_\theta(a_{i,t}|s_{i,t}) \hat{A}_{i,t}^{\text{WS}}\right] \\
J^*(\theta) &= \mathbb{E}_{q,\{\tau_i\}}\left[\frac{1}{G}\sum_{i=1}^G \frac{1}{|\tau_i|}\sum_{t=1}^{|\tau_i|} \pi_\theta(a_{i,t}|s_{i,t}) \hat{A}_{i,t}^{\text{oracle}}\right]
\end{align}

where the advantages are computed using the GRPO normalization:
\begin{align}
\hat{A}_{i,t}^{\text{WS}} &= \frac{R^{\text{WS}}_i - \bar{R}^{\text{WS}}}{\sigma^{\text{WS}}} \\
\hat{A}_{i,t}^{\text{oracle}} &= \frac{R^{\text{oracle}}_i - \bar{R}^{\text{oracle}}}{\sigma^{\text{oracle}}}
\end{align}

with group statistics $\bar{R} = \frac{1}{G}\sum_{i=1}^G R_i$ and $\sigma = \sqrt{\frac{1}{G}\sum_{i=1}^G (R_i - \bar{R})^2}$.

We begin by analyzing the reward decomposition.
The WS-GRPO reward combines preference and final outcome components:
\begin{equation}
R^{\text{WS}}_i = \lambda R^{\text{pref}}_i +  R^{\text{final}}_i
\end{equation}

For each trajectory $\tau_i$, the preference reward is computed as:
\begin{equation}
R^{\text{pref}}_i = \sum_{t=2}^{|\tau_i|} \sigma(P_{\hat{\omega}_n}(h_q, E(s_{i,1:t-1}), E(s_{i,1:t})))
\end{equation}

where $P_{\hat{\omega}_n}(h_q, h_{\text{short}}, h_{\text{long}})$ outputs a preference score for the longer trajectory segment.

The oracle reward uses the true preference model $P_{\omega^*}$:
\begin{equation}
R^{\text{oracle}}_i = \lambda R^{\text{oracle-pref}}_i +  R^{\text{final}}_i
\end{equation}

where:
\begin{equation}
R^{\text{oracle-pref}}_i = \sum_{t=2}^{|\tau_i|} \sigma(P_{\omega^*}(h_q, E(s_{i,1:t-1}), E(s_{i,1:t})))
\end{equation}

Using the bounded error assumption $\epsilon_{\text{pref}} = \left\|P_{\hat{\omega}_n} - P_{\omega^*}\right\|_{\infty}$ and the Lipschitz property of the sigmoid function, we can bound the preference reward error. The sigmoid function $\sigma(z) = \frac{1}{1+e^{-z}}$ has derivative $\sigma'(z) = \sigma(z)(1-\sigma(z)) \leq \frac{1}{4}$, making it $\frac{1}{4}$-Lipschitz.

For each step-wise preference reward:
\begin{align}
&|\sigma(P_{\hat{\omega}_n}(h_q, E(s_{i,1:t-1}), E(s_{i,1:t}))) - \sigma(P_{\omega^*}(h_q, E(s_{i,1:t-1}), E(s_{i,1:t})))| \\
&\leq \frac{1}{4} |P_{\hat{\omega}_n}(h_q, E(s_{i,1:t-1}), E(s_{i,1:t})) - P_{\omega^*}(h_q, E(s_{i,1:t-1}), E(s_{i,1:t}))| \\
&\leq \frac{1}{4} \epsilon_{\text{pref}}
\end{align}

Summing over all steps in trajectory $\tau_i$:
\begin{align}
|R^{\text{pref}}_i - R^{\text{oracle-pref}}_i| &= \left|\sum_{t=2}^{|\tau_i|} \left[\sigma(P_{\hat{\omega}_n}(\cdot)) - \sigma(P_{\omega^*}(\cdot))\right]\right| \\
&\leq \sum_{t=2}^{|\tau_i|} |\sigma(P_{\hat{\omega}_n}(\cdot)) - \sigma(P_{\omega^*}(\cdot))| \\
&\leq \sum_{t=2}^{|\tau_i|} \frac{1}{4} \epsilon_{\text{pref}} \\
&= \frac{|\tau_i| - 1}{4} \epsilon_{\text{pref}} \\
&\leq \frac{T_{\max}}{4} \epsilon_{\text{pref}}
\end{align}

where $T_{\max}$ is the maximum trajectory length.

\begin{align}
|R^{\text{WS}}_i - R^{\text{oracle}}_i| &= |\lambda R^{\text{pref}}_i + R^{\text{final}}_i - \lambda R^{\text{oracle-pref}}_i -  R^{\text{final}}_i| \\
&= |\lambda (R^{\text{pref}}_i - R^{\text{oracle-pref}}_i)| \\
&= \lambda |R^{\text{pref}}_i - R^{\text{oracle-pref}}_i| \\
&\leq \lambda \frac{T_{\max}}{4} \epsilon_{\text{pref}}
\end{align}

The advantage functions are computed using group normalization. For the group statistics:
\begin{align}
|\bar{R}^{\text{WS}} - \bar{R}^{\text{oracle}}| &= \left|\frac{1}{G}\sum_{i=1}^G R^{\text{WS}}_i - \frac{1}{G}\sum_{i=1}^G R^{\text{oracle}}_i\right| \\
&= \frac{1}{G}\left|\sum_{i=1}^G (R^{\text{WS}}_i - R^{\text{oracle}}_i)\right| \\
&\leq \frac{1}{G}\sum_{i=1}^G |R^{\text{WS}}_i - R^{\text{oracle}}_i| \\
&\leq \frac{1}{G} \cdot G \cdot \lambda \frac{T_{\max}}{4} \epsilon_{\text{pref}} \\
&= \lambda \frac{T_{\max}}{4} \epsilon_{\text{pref}}
\end{align}

For the standard deviations, assuming bounded rewards and using the fact that standard deviation is Lipschitz with constant 1:
\begin{equation}
|\sigma^{\text{WS}} - \sigma^{\text{oracle}}| \leq \lambda \frac{T_{\max}}{4} \epsilon_{\text{pref}}
\end{equation}

The advantage difference can be bounded as:
\begin{align}
|\hat{A}_{i,t}^{\text{WS}} - \hat{A}_{i,t}^{\text{oracle}}| &= \left|\frac{R^{\text{WS}}_i - \bar{R}^{\text{WS}}}{\sigma^{\text{WS}}} - \frac{R^{\text{oracle}}_i - \bar{R}^{\text{oracle}}}{\sigma^{\text{oracle}}}\right| \\
&\leq \frac{|R^{\text{WS}}_i - R^{\text{oracle}}_i|}{\min(\sigma^{\text{WS}}, \sigma^{\text{oracle}})} + \frac{|\bar{R}^{\text{WS}} - \bar{R}^{\text{oracle}}|}{\min(\sigma^{\text{WS}}, \sigma^{\text{oracle}})} \\
&\quad + \frac{|R^{\text{oracle}}_i - \bar{R}^{\text{oracle}}| \cdot |\sigma^{\text{WS}} - \sigma^{\text{oracle}}|}{(\sigma^{\text{WS}})(\sigma^{\text{oracle}})}
\end{align}

Assuming the group standard deviations are bounded away from zero (i.e., $\sigma^{\text{WS}}, \sigma^{\text{oracle}} \geq \sigma_{\min} > 0$), we get:
\begin{equation}
|\hat{A}_{i,t}^{\text{WS}} - \hat{A}_{i,t}^{\text{oracle}}| \leq C \lambda \frac{T_{\max}}{4} \epsilon_{\text{pref}}
\end{equation}

for some constant $C > 0$ depending on $\sigma_{\min}$ and reward bounds.

Since the policy class is uniformly bounded, there exists $M > 0$ such that $|\pi_\theta(a|s)| \leq M$ for all $\theta, a, s$. The objective difference is:
\begin{align}
&|\mathbb{E}[J_{\text{WS-GRPO}}(\theta)] - \mathbb{E}[J^*(\theta)]| \\
&= \left|\mathbb{E}_{q,\{\tau_i\}}\left[\frac{1}{G}\sum_{i=1}^G \frac{1}{|\tau_i|}\sum_{t=1}^{|\tau_i|} \pi_\theta(a_{i,t}|s_{i,t}) (\hat{A}_{i,t}^{\text{WS}} - \hat{A}_{i,t}^{\text{oracle}})\right]\right| \\
&\leq \mathbb{E}_{q,\{\tau_i\}}\left[\frac{1}{G}\sum_{i=1}^G \frac{1}{|\tau_i|}\sum_{t=1}^{|\tau_i|} |\pi_\theta(a_{i,t}|s_{i,t})| \cdot |\hat{A}_{i,t}^{\text{WS}} - \hat{A}_{i,t}^{\text{oracle}}|\right] \\
&\leq M \cdot \mathbb{E}_{q,\{\tau_i\}}\left[\frac{1}{G}\sum_{i=1}^G \frac{1}{|\tau_i|}\sum_{t=1}^{|\tau_i|} |\hat{A}_{i,t}^{\text{WS}} - \hat{A}_{i,t}^{\text{oracle}}|\right] \\
&\leq M \cdot C \lambda \frac{T_{\max}}{4} \epsilon_{\text{pref}}
\end{align}

Absorbing the constants $M$ and $C$ into a single constant, we obtain:
\begin{equation}
\left|\mathbb{E}[J_{\text{WS-GRPO}}(\theta)] - \mathbb{E}[J^*(\theta)]\right| \leq \frac{\lambda T_{\max}}{4} \cdot \epsilon_{\text{pref}}
\end{equation}

This bound holds with probability at least $1-\delta$ when $\epsilon_{\text{pref}}$ is the bound from Theorem~\ref{app:preference-consistency}.

\end{theorem}

\begin{theorem}[WS-GRPO Generalization Bound]
\label{app:ws-grpo-generalization}

Now we establish the comprehensive generalization bound for WS-GRPO by combining all error sources through a union bound.
We decompose the generalization error into three components.

Let $\mathcal{R}(\pi_\theta)$ denote the true risk (expected performance) and $\hat{\mathcal{R}}(\pi_\theta)$ denote the empirical risk computed on the training set of size $n$. We want to bound $\mathcal{R}(\pi_\theta) - \hat{\mathcal{R}}(\pi_\theta)$.

For WS-GRPO, the empirical risk involves both policy gradient terms and preference reward terms:
\begin{equation}
\hat{\mathcal{R}}(\pi_\theta) = \frac{1}{n}\sum_{j=1}^n \left[\frac{1}{G}\sum_{i=1}^G \frac{1}{|\tau_{j,i}|}\sum_{t=1}^{|\tau_{j,i}|} \log \pi_\theta(a_{j,i,t}|s_{j,i,t}) \hat{A}_{j,i,t}^{\text{WS}}\right]
\end{equation}

where $\hat{A}_{j,i,t}^{\text{WS}}$ are advantages computed using WS-GRPO rewards.

For the policy class $\mathcal{H}$ with VC-dimension $d$, the Rademacher complexity of the policy class is:
\begin{equation}
\mathfrak{R}_n(\mathcal{H}) = \mathbb{E}_{\boldsymbol{\sigma}}\left[\sup_{\pi \in \mathcal{H}} \frac{1}{n}\sum_{j=1}^n \sigma_j \ell(\pi, x_j)\right] \leq \sqrt{\frac{2d\log(2en/d)}{n}}
\end{equation}

where $\boldsymbol{\sigma} = (\sigma_1, \ldots, \sigma_n)$ are independent Rademacher variables and $\ell(\pi, x_j)$ represents the loss for policy $\pi$ on example $x_j$.

Using McDiarmid's inequality with the bounded difference assumption (policy outputs are bounded), we have:
\begin{align}
&\mathbb{P}\left[\sup_{\pi \in \mathcal{H}} \left|\mathcal{R}_{\text{GRPO}}(\pi) - \hat{\mathcal{R}}_{\text{GRPO}}(\pi)\right| \geq 2\mathfrak{R}_n(\mathcal{H}) + \sqrt{\frac{2\log(2/\delta_1)}{n}}\right] \leq \delta_1
\end{align}

Substituting the Rademacher complexity bound:
\begin{align}
&\mathbb{P}\left[\sup_{\pi \in \mathcal{H}} \left|\mathcal{R}_{\text{GRPO}}(\pi) - \hat{\mathcal{R}}_{\text{GRPO}}(\pi)\right| \geq 2\sqrt{\frac{2d\log(2en/d)}{n}} + \sqrt{\frac{2\log(2/\delta_1)}{n}}\right] \leq \delta_1
\end{align}

Using the inequality $\sqrt{a} + \sqrt{b} \leq \sqrt{2(a+b)}$ for $a,b \geq 0$:
\begin{equation}
\mathbb{P}\left[\sup_{\pi \in \mathcal{H}} \left|\mathcal{R}_{\text{GRPO}}(\pi) - \hat{\mathcal{R}}_{\text{GRPO}}(\pi)\right| \geq \sqrt{\frac{8d\log(2en/d) + 8\log(2/\delta_1)}{n}}\right] \leq \delta_1
\end{equation}

Each preference reward is bounded:
\begin{equation}
|R^{\text{pref}}_i| = \left|\sum_{t=2}^{|\tau_i|} \sigma(P_{\hat{\omega}_n}(\cdot))\right| \leq \sum_{t=2}^{|\tau_i|} 1 = |\tau_i| - 1 \leq T_{\max}
\end{equation}

Since the preference model output is bounded by $|P_{\hat{\omega}_n}(\cdot)| \leq B$, and $\sigma(z) \in [0,1]$, we have:
\begin{equation}
|R^{\text{pref}}_i| \leq BT_{\max}
\end{equation}

The preference-augmented loss function is:
\begin{equation}
\ell_{\text{pref}}(\pi, q, \{\tau_i\}) = \frac{1}{G}\sum_{i=1}^G \frac{\lambda R^{\text{pref}}_i}{|\tau_i|}\sum_{t=1}^{|\tau_i|} \log \pi(a_{i,t}|s_{i,t})
\end{equation}

Since $|\log \pi(a|s)| \leq \log(1/\pi_{\min}) \leq L_{\pi}$ for some constant $L_{\pi}$, the preference loss is bounded by:
\begin{equation}
|\ell_{\text{pref}}(\pi, q, \{\tau_i\})| \leq \frac{1}{G}\sum_{i=1}^G \frac{\lambda BT_{\max}}{|\tau_i|} \cdot |\tau_i| \cdot L_{\pi} = \lambda BT_{\max} L_{\pi}
\end{equation}

Applying Hoeffding's inequality to the bounded preference rewards:
\begin{equation}
\mathbb{P}\left[\left|\mathbb{E}[\ell_{\text{pref}}] - \hat{\mathbb{E}}[\ell_{\text{pref}}]\right| \geq t\right] \leq 2\exp\left(-\frac{2nt^2}{(\lambda BT_{\max} L_{\pi})^2}\right)
\end{equation}

Setting the right-hand side equal to $\delta_2$ and solving for $t$:
\begin{equation}
t = \lambda BT_{\max} L_{\pi} \sqrt{\frac{\log(2/\delta_2)}{2n}}
\end{equation}

Absorbing $L_{\pi}$ into the bound and using a looser but cleaner bound:
\begin{equation}
\mathbb{P}\left[\left|\mathbb{E}[R^{\text{pref}}] - \hat{\mathbb{E}}[R^{\text{pref}}]\right| \geq \lambda BT_{\max}\sqrt{\frac{2\log(2/\delta_2)}{n}}\right] \leq \delta_2
\end{equation}

From Theorems~\ref{app:preference-consistency} and~\ref{app:policy-robustness}, the preference model error contributes an additional term. The preference model error is bounded by:
\begin{equation}
\epsilon_{\text{pref}} = \left\|P_{\hat{\omega}_n} - P_{\omega^*}\right\|_{\infty} \leq \sqrt{\frac{2d_P \log(2en/d_P) + 2\log(2/\delta_3)}{n}}
\end{equation}

This error propagates to the policy objective with the bound from Theorem~\ref{app:policy-robustness}:
\begin{equation}
\left|\mathbb{E}[J_{\text{WS-GRPO}}(\theta)] - \mathbb{E}[J^{*}(\theta)]\right| \leq \frac{\lambda T_{\max}}{4} \epsilon_{\text{pref}}
\end{equation}

Substituting the preference model error bound:
\begin{align}
\left|\mathbb{E}[J_{\text{WS-GRPO}}(\theta)] - \mathbb{E}[J^{*}(\theta)]\right| &\leq \frac{\lambda T_{\max}}{4} \sqrt{\frac{2d_P \log(2en/d_P) + 2\log(2/\delta_3)}{n}}
\end{align}

This gives us:
\begin{equation}
\mathbb{P}\left[\left|\mathbb{E}[J_{\text{WS-GRPO}}(\theta)] - \mathbb{E}[J^{*}(\theta)]\right| \geq \frac{\lambda T_{\max}}{4} \sqrt{\frac{2d_P \log(2en/d_P) + 2\log(2/\delta_3)}{n}}\right] \leq \delta_3
\end{equation}

Now we combine all error sources using the union bound.
Setting $\delta_1 = \delta_2 = \delta_3 = \delta/3$ and applying the union bound, with probability at least $1-\delta$:

\begin{align}
\mathcal{R}(\pi_\theta) - \hat{\mathcal{R}}(\pi_\theta) &\leq |\mathcal{R}_{\text{GRPO}}(\pi_\theta) - \hat{\mathcal{R}}_{\text{GRPO}}(\pi_\theta)| \\
&\quad + |\mathbb{E}[\ell_{\text{pref}}] - \hat{\mathbb{E}}[\ell_{\text{pref}}]| \\
&\quad + |\mathbb{E}[J_{\text{WS-GRPO}}(\theta)] - \mathbb{E}[J^{*}(\theta)]|
\end{align}

Substituting the individual bounds:
\begin{align}
\mathcal{R}(\pi_\theta) - \hat{\mathcal{R}}(\pi_\theta) &\leq \sqrt{\frac{8d\log(2en/d) + 8\log(6/\delta)}{n}} \\
&\quad + \lambda BT_{\max}\sqrt{\frac{2\log(6/\delta)}{n}} \\
&\quad + \frac{\lambda T_{\max}}{4} \sqrt{\frac{2d_P \log(2en/d_P) + 2\log(6/\delta)}{n}}
\end{align}

To obtain a more compact form, we combine these three terms. Let $d_{\max} = \max(d, d_P)$ and observe that all terms have the same $O(\sqrt{\log n/n})$ rate. Using the inequality $\sqrt{a} + \sqrt{b} + \sqrt{c} \leq \sqrt{3(a+b+c)}$ and factoring out common terms:

\begin{align}
\mathcal{R}(\pi_\theta) - \hat{\mathcal{R}}(\pi_\theta) &\leq \sqrt{\frac{8d\log(2en/d) + 8\log(12/\delta)}{n}} \\
&\quad + \lambda BT_{\max}\sqrt{\frac{2\log(12/\delta)}{n}} \\
&\quad + \frac{\lambda T_{\max}}{4} \sqrt{\frac{2d_P \log(2en/d_P) + 2\log(12/\delta)}{n}} \\
&\leq \sqrt{\frac{C_1 d_{\max}\log(en/d_{\max}) + C_2 \lambda^2 (BT_{\max})^2 + C_3 \log(1/\delta)}{n}}
\end{align}

where $C_1, C_2, C_3 > 0$ are universal constants that absorb the numerical factors. This compact form shows that the generalization error scales as:
\begin{equation}
\mathcal{R}(\pi_\theta) - \hat{\mathcal{R}}(\pi_\theta) = \tilde{O}\left(\sqrt{\frac{d_{\max} + \lambda^2 (BT_{\max})^2}{n}}\right)
\end{equation}

where $\tilde{O}$ hides logarithmic factors in $n$ and $\delta$. This demonstrates that WS-GRPO maintains the standard statistical learning rate while the preference-specific terms contribute additively to the complexity, controlled by the mixing weight $\lambda$ and model capacities.

\end{theorem}

\begin{tcolorbox}[
    title={Prompt~\refstepcounter{prompt}\theprompt: AI2-ARC Scientific Reasoning},
    colback=white,
    colframe=black,
    colbacktitle=black,
    coltitle=white,
    fonttitle=\bfseries,
    width=\textwidth,
    sharp corners,
    boxrule=0.5pt,
    before skip=10pt,
    after skip=10pt
]

\textbf{System Prompt:}\\
A conversation between User and Assistant. The User asks a question, and the Assistant solves it. The Assistant first thinks about the reasoning process in the mind and then provides the User with the answer. The reasoning process is enclosed within \texttt{<think> </think>} and answer is enclosed within \texttt{<answer> </answer>} tags, respectively, i.e., \texttt{<think>} reasoning process here \texttt{</think>} \texttt{<answer>} answer here\texttt{</answer>}. \texttt{<answer>} must contain only the letter of your choice (A, B, C, D).

\vspace{0.75em}

\textbf{User Prompt:}
\begin{verbatim}
<Question>
<Options>
\end{verbatim}
\label{box:ai2-arc}
\end{tcolorbox}

\begin{tcolorbox}[
    title={Prompt~\refstepcounter{prompt}\theprompt: CommonsenseQA Reasoning},
    colback=white,
    colframe=black,
    colbacktitle=black,
    coltitle=white,
    fonttitle=\bfseries,
    width=\textwidth,
    sharp corners,
    boxrule=0.5pt,
    before skip=10pt,
    after skip=10pt
]

\textbf{System Prompt:}\\
A conversation between User and Assistant. The User asks a question, and the Assistant solves it. The Assistant first thinks about the reasoning process in the mind and then provides the User with the answer. The reasoning process is enclosed within \texttt{<think> </think>} and answer is enclosed within \texttt{<answer> </answer>} tags, respectively, i.e., \texttt{<think>} reasoning process here \texttt{</think>} \texttt{<answer>} answer here\texttt{</answer>}. \texttt{<answer>} must contain only the letter of your choice (A, B, C, D, or E).

\vspace{0.75em}

\textbf{User Prompt:}
\begin{verbatim}
<Question>
<Options>
\end{verbatim}
\label{box:commonsenseqa}
\end{tcolorbox}

\subsection{Dataset Details}

\begin{figure}[h]
    \centering
    \includegraphics[width=\linewidth]{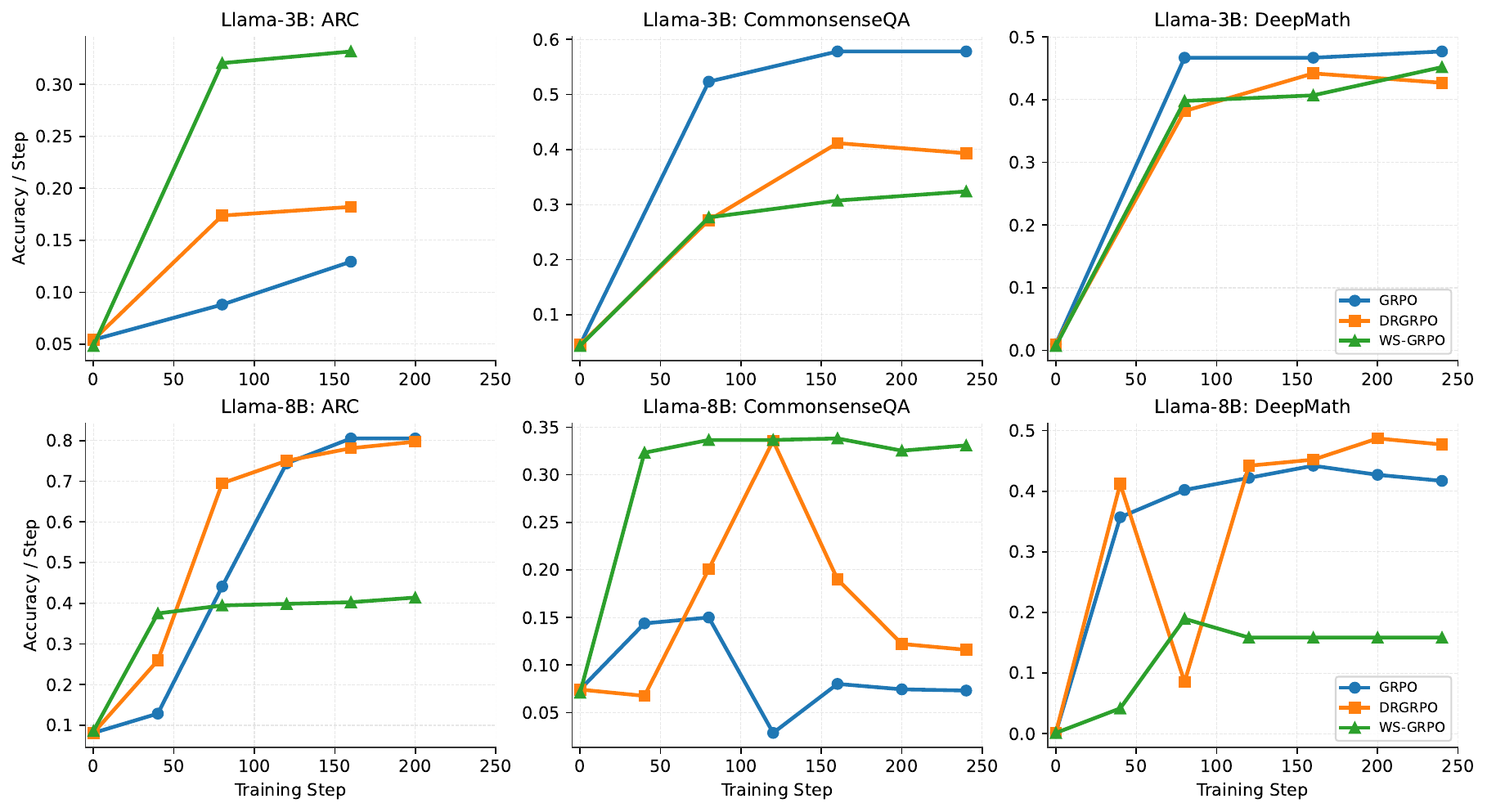}
    \caption{Validation step-efficiency (Pass@1 / average reasoning steps) across training for Llama models. Higher values indicate greater accuracy per reasoning step.}
    \label{fig:llama_step_efficiency}
\end{figure}

\begin{table}[tbp]
\centering
\small
\setlength{\tabcolsep}{8pt}
\renewcommand{\arraystretch}{1.15}

\begin{tabular}{l|ccc}
\toprule
Dataset & Training & Validation & Testing \\
\midrule
ARC & 1813 & 129 & 648 \\
CommonsenseQA  & 2000 & 200 & 2740 \\
Deepmath & 2000 & 200 & 2883 \\
GSM8K & 2000 & 200 & 2198 \\

\bottomrule
\end{tabular}
\caption{Training/Validation/Testing Split for ARC and CommonsenseQA datasets.}
\label{tab:split}
\end{table}

\definecolor{wsGreen}{RGB}{226,245,231}   %
\definecolor{impGreen}{RGB}{214,238,214}  %
\definecolor{headerGray}{RGB}{248,248,248}

\begin{table*}[t]
\centering
\caption{\textbf{Main results (accuracy + efficiency) for GRPO vs DRGRPO vs WS-GRPO.}
We report \textbf{test-set} Pass@1 accuracy (↑) and \textbf{test-set} average reasoning steps (↓).
Efficiency is additionally reported as the \textbf{mean completion length} in tokens (↓) measured during evaluation at the \textbf{final validation checkpoint}.
\textbf{Improve} compares WS-GRPO against the best baseline among \{GRPO, DRGRPO\} for each metric
(max for Pass@1; min for length/steps). Green cells indicate improvements in cost metrics (lower steps/length).}
\label{tab:llama_efficiency}
\renewcommand{\arraystretch}{1.12}
\setlength{\tabcolsep}{5.0pt}
\resizebox{\textwidth}{!}{%
\begin{tabular}{ll|cccc|cccc}
\toprule
\rowcolor{headerGray}
& & \multicolumn{4}{c|}{\textbf{Llama-3.2-3B-Instruct}} & \multicolumn{4}{c}{\textbf{Llama-3.1-8B-Instruct}} \\
\rowcolor{headerGray}
\cmidrule(lr){3-6}\cmidrule(lr){7-10}
\cellcolor{white}\textbf{Dataset} & \cellcolor{white}\textbf{Metric} &
\mth{\textbf{GRPO}} & \mth{\textbf{DRGRPO}} & \mth{\textbf{WS-GRPO}} & \textbf{Improve} &
\mth{\textbf{GRPO}} & \mth{\textbf{DRGRPO}} & \mth{\textbf{WS-GRPO}} & \textbf{Improve} \\
\midrule

\multirow{3}{*}{ARC}
& Pass@1 $\uparrow$
  & 0.794 & 0.782 & \textbf{0.748} & $\Delta=-0.046$
  & 0.816 & 0.841 & \textbf{0.824} & $\Delta=-0.017$ \\
& Eval length (tok.) $\downarrow$
  & 137.20 & 97.91 & \cellcolor{wsGreen}\textbf{14.00} & \good{$\downarrow 85.7\%$}
  & 13.05 & 8.00 & \textbf{20.00} & $\uparrow 150\%$ \\
& Avg steps $\downarrow$
  & 5.07 & 4.22 & \cellcolor{wsGreen}\textbf{2.04} & \good{$\downarrow 51.7\%$}
  & 1.00 & 1.00 & \textbf{2.00} & $\uparrow 100\%$ \\
\midrule

\multirow{3}{*}{CommonsenseQA}
& Pass@1 $\uparrow$
  & 0.714 & 0.717 & \textbf{0.688} & $\Delta=-0.029$
  & 0.737 & 0.749 & \textbf{0.721} & $\Delta=-0.028$ \\
& Eval length (tok.) $\downarrow$
  & 38.20 & 51.60 & \cellcolor{wsGreen}\textbf{37.49} & \good{$\downarrow 1.9\%$}
  & 165.03 & 441.09 & \cellcolor{wsGreen}\textbf{38.44} & \good{$\downarrow 76.7\%$} \\
& Avg steps $\downarrow$
  & 1.17 & 1.80 & \textbf{2.01} & $\uparrow 71.8\%$
  & 9.36 & 5.81 & \cellcolor{wsGreen}\textbf{2.11} & \good{$\downarrow 63.7\%$} \\
\midrule

\multirow{3}{*}{DeepMath}
& Pass@1 $\uparrow$
  & 0.455 & 0.460 & \textbf{0.464} & $\Delta=+0.004$
  & 0.464 & 0.463 & \textbf{0.372} & $\Delta=-0.092$ \\
& Eval length (tok.) $\downarrow$
  & 9.02 & 8.14 & \textbf{12.66} & $\uparrow 55.5\%$
  & 9.09 & 512.00 & \cellcolor{wsGreen}\textbf{41.49} & $\uparrow 356\%$ \\
& Avg steps $\downarrow$
  & 1.00 & 1.01 & \cellcolor{wsGreen}\textbf{1.00} & \good{$\downarrow 0\%$}
  & 1.00 & 1.00 & \textbf{2.00} & $\uparrow 100\%$ \\
\bottomrule
\end{tabular}%
}
\end{table*}

\subsection{Length Penalty}
\label{sec:length_penalty}
To encourage reasoning trajectories with appropriate length, we apply a length penalty to the stepwise reward:
\begin{equation}
\ell(n) = 
\begin{cases}
-\alpha (n_{\min} - n) & \text{if } n < n_{\min} \\
0 & \text{if } n_{\min} \leq n \leq n_{\max} \\
-\alpha (n - n_{\max}) & \text{if } n > n_{\max}
\end{cases}
\end{equation}
where $n$ is the number of reasoning steps, $n_{\min}=3$, $n_{\max}=6$, and $\alpha=0.1$. The adjusted stepwise reward is computed as:
\begin{equation}
r_{\text{step}} = \frac{\bar{p} + \ell(n)}{n}
\end{equation}
where $\bar{p}$ is the mean preference probability across steps. This normalization ensures fair comparison across trajectories of different lengths.

\subsection{Training Hyperparameters}
\label{sec:training_hyperparameters}
\textbf{Phase I - Preference Learning:}
We use Qwen2.5-3B-Instruct to generate 4 reasoning trajectories per question, yielding an average of 85,425 preference pairs per dataset (ranging from 60K to 103K). We apply a 90\%/10\% train-validation split. The preference model uses a frozen FLAN-T5 encoder followed by a lightweight MLP classifier with hidden dimension 512, with batch size 32 and learning rate $5\times10^{-5}$.

\textbf{Phase II - Policy Optimization:}
We use $G=8$ rollouts per prompt with learning rate $\eta=1\times10^{-5}$. The mixing weight is set to $\lambda=0.1$ to balance preference rewards with outcome correctness. Length penalty coefficient $\alpha=0.1$ encourages trajectories between 3 and 6 reasoning steps (see \Cref{sec:length_penalty} for details). Step-wise rewards are normalized by trajectory length to prevent bias toward longer sequences.

\subsection{Ability to anticipate final correctness for preference model}

\begin{figure}[h]
    \centering
    \includegraphics[width=0.5\linewidth]{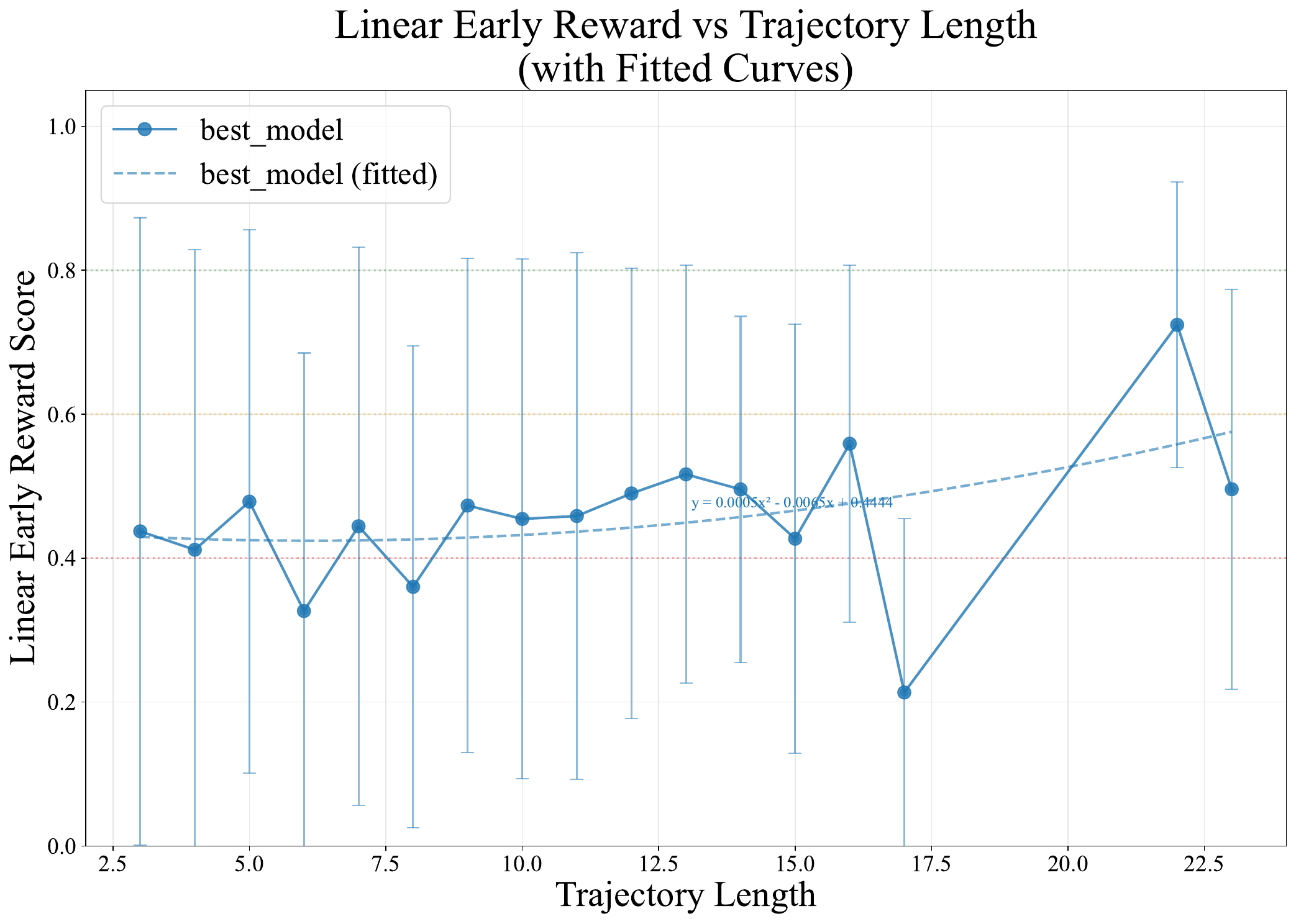}
    \caption{Preference Model Effectiveness}
    \label{fig:module_reliability}
\end{figure}

We further designed a fine-grained metric to assess the model’s ability to anticipate final correctness from partial trajectories. For each pair of correct and incorrect trajectories of length $n$, we evaluate all prefixes of length $2, 3, \dots, n$. A correctly ordered prefix pair receives a weight inversely proportional to its step index, emphasizing early correct predictions. The cumulative score is normalized by the maximum achievable value. Figure~\ref{fig:module_reliability} shows that this metric increases with trajectory length, confirming that the model not only distinguishes full trajectories but also reliably ranks partial reasoning paths. The upward trend suggests that longer trajectories provide more discriminative signal, enabling more accurate stepwise reward assignment during policy optimization.

%% file: icml2026_main.bbl
\begin{thebibliography}{61}
\providecommand{\natexlab}[1]{#1}
\providecommand{\url}[1]{\texttt{#1}}
\expandafter\ifx\csname urlstyle\endcsname\relax
  \providecommand{\doi}[1]{doi: #1}\else
  \providecommand{\doi}{doi: \begingroup \urlstyle{rm}\Url}\fi

\bibitem[Aggarwal \& Welleck(2025)Aggarwal and Welleck]{aggarwal2025l1}
Aggarwal, P. and Welleck, S.
\newblock L1: Controlling how long a reasoning model thinks with reinforcement learning, 2025.
\newblock URL \url{https://arxiv.org/abs/2503.04697}.

\bibitem[Azizi et~al.(2025)Azizi, Potraghloo, and Pedram]{azizi2025activation}
Azizi, S., Potraghloo, E.~B., and Pedram, M.
\newblock Activation steering for chain-of-thought compression, 2025.
\newblock URL \url{https://arxiv.org/abs/2507.04742}.

\bibitem[Bartlett \& Mendelson(2002)Bartlett and Mendelson]{bartlett2002rademacher}
Bartlett, P.~L. and Mendelson, S.
\newblock Rademacher and gaussian complexities: Risk bounds and structural results.
\newblock \emph{Journal of machine learning research}, 3\penalty0 (Nov):\penalty0 463--482, 2002.

\bibitem[Chen et~al.(2025)Chen, Zhang, Hong, Kundu, and Wang]{chen2025seal}
Chen, R., Zhang, Z., Hong, J., Kundu, S., and Wang, Z.
\newblock Seal: Steerable reasoning calibration of large language models for free, 2025.
\newblock URL \url{https://arxiv.org/abs/2504.07986}.

\bibitem[Clark et~al.(2018)Clark, Cowhey, Etzioni, Khot, Sabharwal, Schoenick, and Tafjord]{clark2018think}
Clark, P., Cowhey, I., Etzioni, O., Khot, T., Sabharwal, A., Schoenick, C., and Tafjord, O.
\newblock Think you have solved question answering? try arc, the ai2 reasoning challenge.
\newblock \emph{arXiv preprint arXiv:1803.05457}, 2018.

\bibitem[Cobbe et~al.(2021)Cobbe, Kosaraju, Bavarian, Chen, Jun, Kaiser, Plappert, Tworek, Hilton, Nakano, Hesse, and Schulman]{cobbe2021gsm8k}
Cobbe, K., Kosaraju, V., Bavarian, M., Chen, M., Jun, H., Kaiser, L., Plappert, M., Tworek, J., Hilton, J., Nakano, R., Hesse, C., and Schulman, J.
\newblock Training verifiers to solve math word problems.
\newblock \emph{arXiv preprint arXiv:2110.14168}, 2021.

\bibitem[Cui et~al.(2025)Cui, Yuan, Wang, Wang, Zhang, Chen, Li, He, Fan, Yu, et~al.]{cui2025process}
Cui, G., Yuan, L., Wang, Z., Wang, H., Zhang, Y., Chen, J., Li, W., He, B., Fan, Y., Yu, T., et~al.
\newblock Process reinforcement through implicit rewards.
\newblock \emph{arXiv preprint arXiv:2502.01456}, 2025.

\bibitem[Dai et~al.(2025)Dai, Yang, and Si]{dai2025sgrpo}
Dai, M., Yang, C., and Si, Q.
\newblock S-grpo: Early exit via reinforcement learning in reasoning models, 2025.
\newblock URL \url{https://arxiv.org/abs/2505.07686}.

\bibitem[Dubois et~al.(2024)Dubois, Galambosi, Liang, and Hashimoto]{dubois2024length}
Dubois, Y., Galambosi, B., Liang, P., and Hashimoto, T.~B.
\newblock Length-controlled alpacaeval: A simple way to debias automatic evaluators.
\newblock \emph{arXiv preprint arXiv:2404.04475}, 2024.

\bibitem[Fei et~al.(2025)Fei, Kong, Liang, Lin, Yang, Tang, Chen, and Hua]{fei2025self}
Fei, W., Kong, H., Liang, S., Lin, Y., Yang, Y., Tang, J., Chen, L., and Hua, X.
\newblock Self-guided process reward optimization with masked step advantage for process reinforcement learning.
\newblock \emph{arXiv preprint arXiv:2507.01551}, 2025.

\bibitem[Grattafiori et~al.(2024)Grattafiori, Dubey, Jauhri, Pandey, Kadian, Al-Dahle, Letman, Mathur, Schelten, Vaughan, et~al.]{grattafiori2024llama}
Grattafiori, A., Dubey, A., Jauhri, A., Pandey, A., Kadian, A., Al-Dahle, A., Letman, A., Mathur, A., Schelten, A., Vaughan, A., et~al.
\newblock The llama 3 herd of models.
\newblock \emph{arXiv preprint arXiv:2407.21783}, 2024.

\bibitem[Guo et~al.(2025)Guo, Yang, Zhang, Song, Zhang, Xu, Zhu, Ma, Wang, Bi, et~al.]{guo2025deepseek}
Guo, D., Yang, D., Zhang, H., Song, J., Zhang, R., Xu, R., Zhu, Q., Ma, S., Wang, P., Bi, X., et~al.
\newblock Deepseek-r1: Incentivizing reasoning capability in llms via reinforcement learning.
\newblock \emph{arXiv preprint arXiv:2501.12948}, 2025.

\bibitem[Guo et~al.(2023)Guo, Jin, Liu, Huang, Shi, Yu, Liu, Li, Xiong, Xiong, et~al.]{guo2023evaluating}
Guo, Z., Jin, R., Liu, C., Huang, Y., Shi, D., Yu, L., Liu, Y., Li, J., Xiong, B., Xiong, D., et~al.
\newblock Evaluating large language models: A comprehensive survey.
\newblock \emph{arXiv preprint arXiv:2310.19736}, 2023.

\bibitem[He et~al.(2025)He, Liang, Xu, Liu, Chen, Wang, Song, Yu, Liang, Wang, et~al.]{he2025deepmath}
He, Z., Liang, T., Xu, J., Liu, Q., Chen, X., Wang, Y., Song, L., Yu, D., Liang, Z., Wang, W., et~al.
\newblock Deepmath-103k: A large-scale, challenging, decontaminated, and verifiable mathematical dataset for advancing reasoning.
\newblock \emph{arXiv preprint arXiv:2504.11456}, 2025.

\bibitem[Hosseini et~al.(2024)Hosseini, Yuan, Malkin, Courville, Sordoni, and Agarwal]{hosseini2024v}
Hosseini, A., Yuan, X., Malkin, N., Courville, A., Sordoni, A., and Agarwal, R.
\newblock V-star: Training verifiers for self-taught reasoners.
\newblock \emph{arXiv preprint arXiv:2402.06457}, 2024.

\bibitem[Hou et~al.(2025)Hou, Zhang, Ji, Liu, Qian, Andreas, and Chang]{hou2025thinkprune}
Hou, B., Zhang, Y., Ji, J., Liu, Y., Qian, K., Andreas, J., and Chang, S.
\newblock Thinkprune: Pruning long chain-of-thought of llms via reinforcement learning, 2025.
\newblock URL \url{https://arxiv.org/abs/2504.01296}.

\bibitem[Huang et~al.(2025{\natexlab{a}})Huang, Wang, Yang, Zhao, Li, Lin, Zhang, Rajmohan, and Zhang]{huang2025lean}
Huang, C., Wang, L., Yang, F., Zhao, P., Li, Z., Lin, Q., Zhang, D., Rajmohan, S., and Zhang, Q.
\newblock Lean and mean: Decoupled value policy optimization with global value guidance.
\newblock \emph{arXiv preprint arXiv:2502.16944}, 2025{\natexlab{a}}.

\bibitem[Huang et~al.(2025{\natexlab{b}})Huang, Huang, Wu, Yu, McAuley, and Yao]{huang2025listwise}
Huang, H., Huang, C., Wu, J., Yu, T., McAuley, J., and Yao, L.
\newblock Listwise preference diffusion optimization for user behavior trajectories prediction.
\newblock \emph{arXiv preprint arXiv:2511.00530}, 2025{\natexlab{b}}.

\bibitem[Huang et~al.(2025{\natexlab{c}})Huang, Chen, Ruan, Zhang, Wei, and Dong]{huang2025mitigat}
Huang, Y., Chen, H., Ruan, S., Zhang, Y., Wei, X., and Dong, Y.
\newblock Mitigating overthinking in large reasoning models via manifold steering, 2025{\natexlab{c}}.
\newblock URL \url{https://arxiv.org/abs/2505.22411}.

\bibitem[Kikuchi et~al.(2016)Kikuchi, Neubig, Sasano, Takamura, and Okumura]{kikuchi2016controlling}
Kikuchi, Y., Neubig, G., Sasano, R., Takamura, H., and Okumura, M.
\newblock Controlling output length in neural encoder-decoders.
\newblock \emph{arXiv preprint arXiv:1609.09552}, 2016.

\bibitem[Lei et~al.(2023)Lei, He, Yuan, and Tao]{lei2023understanding}
Lei, S., He, F., Yuan, Y., and Tao, D.
\newblock Understanding deep learning via decision boundary.
\newblock \emph{IEEE Transactions on Neural Networks and Learning Systems}, 2023.

\bibitem[Li et~al.(2025{\natexlab{a}})Li, Wang, Wu, Surana, Yu, McAuley, and Shang]{li2025importance}
Li, X., Wang, C., Wu, J., Surana, R., Yu, T., McAuley, J., and Shang, J.
\newblock Importance sampling for multi-negative multimodal direct preference optimization.
\newblock \emph{arXiv preprint arXiv:2509.25717}, 2025{\natexlab{a}}.

\bibitem[Li et~al.(2025{\natexlab{b}})Li, Wang, Zhu, Zhao, Lu, She, and Zhang]{li2025branchgrpo}
Li, Y., Wang, Y., Zhu, Y., Zhao, Z., Lu, M., She, Q., and Zhang, S.
\newblock Branchgrpo: Stable and efficient grpo with structured branching in diffusion models.
\newblock \emph{arXiv preprint arXiv:2509.06040}, 2025{\natexlab{b}}.

\bibitem[Lightman et~al.(2023)Lightman, Kosaraju, Burda, Edwards, Baker, Lee, Leike, Schulman, Sutskever, and Cobbe]{lightman2023let}
Lightman, H., Kosaraju, V., Burda, Y., Edwards, H., Baker, B., Lee, T., Leike, J., Schulman, J., Sutskever, I., and Cobbe, K.
\newblock Let's verify step by step.
\newblock In \emph{The Twelfth International Conference on Learning Representations}, 2023.

\bibitem[Liu et~al.(2025{\natexlab{a}})Liu, Cao, Ren, Zhou, Dong, Ma, Han, and Zhang]{liu2025bingo}
Liu, H., Cao, L., Ren, Y., Zhou, M., Dong, H., Ma, X., Han, S., and Zhang, D.
\newblock Bingo: Boosting efficient reasoning of llms via dynamic and significance-based reinforcement learning, 2025{\natexlab{a}}.
\newblock URL \url{https://arxiv.org/abs/2506.08125}.

\bibitem[Liu et~al.(2025{\natexlab{b}})Liu, Zhou, Deng, Huang, Liu, Deng, Zhang, and He]{liu2025laser}
Liu, W., Zhou, R., Deng, Y., Huang, Y., Liu, J., Deng, Y., Zhang, Y., and He, J.
\newblock Learn to reason efficiently with adaptive length-based reward shaping, 2025{\natexlab{b}}.
\newblock URL \url{https://arxiv.org/abs/2505.15612}.

\bibitem[Liu et~al.(2025{\natexlab{c}})Liu, Chen, Li, Qi, Pang, Du, Lee, and Lin]{liu2025understanding}
Liu, Z., Chen, C., Li, W., Qi, P., Pang, T., Du, C., Lee, W.~S., and Lin, M.
\newblock Understanding r1-zero-like training: A critical perspective.
\newblock \emph{arXiv preprint arXiv:2503.20783}, 2025{\natexlab{c}}.

\bibitem[Madaan et~al.(2023)Madaan, Tandon, Gupta, Hallinan, Gao, Wiegreffe, Alon, Dziri, Prabhumoye, Yang, et~al.]{madaan2023self}
Madaan, A., Tandon, N., Gupta, P., Hallinan, S., Gao, L., Wiegreffe, S., Alon, U., Dziri, N., Prabhumoye, S., Yang, Y., et~al.
\newblock Self-refine: Iterative refinement with self-feedback.
\newblock \emph{Advances in Neural Information Processing Systems}, 36:\penalty0 46534--46594, 2023.

\bibitem[Miao et~al.(2023)Miao, Teh, and Rainforth]{miao2023selfcheck}
Miao, N., Teh, Y.~W., and Rainforth, T.
\newblock Selfcheck: Using llms to zero-shot check their own step-by-step reasoning.
\newblock \emph{arXiv preprint arXiv:2308.00436}, 2023.

\bibitem[Mohri et~al.(2018)Mohri, Rostamizadeh, and Talwalkar]{mohri2018foundations}
Mohri, M., Rostamizadeh, A., and Talwalkar, A.
\newblock \emph{Foundations of machine learning}.
\newblock MIT press, 2018.

\bibitem[Mroueh(2025)]{mroueh2025reinforcement}
Mroueh, Y.
\newblock Reinforcement learning with verifiable rewards: Grpo's effective loss, dynamics, and success amplification.
\newblock \emph{arXiv preprint arXiv:2503.06639}, 2025.

\bibitem[Murray \& Chiang(2018)Murray and Chiang]{murray2018correcting}
Murray, K. and Chiang, D.
\newblock Correcting length bias in neural machine translation.
\newblock \emph{arXiv preprint arXiv:1808.10006}, 2018.

\bibitem[Ning et~al.(2025)Ning, Li, Fang, Tan, and Liu]{ning2025longshort}
Ning, Y., Li, W., Fang, J., Tan, N., and Liu, H.
\newblock Not all thoughts are generated equal: Efficient llm reasoning via multi-turn reinforcement learning, 2025.
\newblock URL \url{https://arxiv.org/abs/2505.11827}.

\bibitem[Pang \& Jin(2025)Pang and Jin]{pang2025theory}
Pang, L. and Jin, R.
\newblock On the theory and practice of grpo: A trajectory-corrected approach with fast convergence.
\newblock \emph{arXiv preprint arXiv:2508.02833}, 2025.

\bibitem[Rafailov et~al.(2023)Rafailov, Sharma, Mitchell, Manning, Ermon, and Finn]{rafailov2023direct}
Rafailov, R., Sharma, A., Mitchell, E., Manning, C.~D., Ermon, S., and Finn, C.
\newblock Direct preference optimization: Your language model is secretly a reward model.
\newblock \emph{Advances in neural information processing systems}, 36:\penalty0 53728--53741, 2023.

\bibitem[Saito et~al.(2023)Saito, Wachi, Wataoka, and Akimoto]{saito2023verbosity}
Saito, K., Wachi, A., Wataoka, K., and Akimoto, Y.
\newblock Verbosity bias in preference labeling by large language models.
\newblock \emph{arXiv preprint arXiv:2310.10076}, 2023.

\bibitem[Shao et~al.(2024)Shao, Wang, Zhu, Xu, Song, Bi, Zhang, Zhang, Li, Wu, et~al.]{shao2024deepseekmath}
Shao, Z., Wang, P., Zhu, Q., Xu, R., Song, J., Bi, X., Zhang, H., Zhang, M., Li, Y., Wu, Y., et~al.
\newblock Deepseekmath: Pushing the limits of mathematical reasoning in open language models.
\newblock \emph{arXiv preprint arXiv:2402.03300}, 2024.

\bibitem[Shen et~al.(2026)Shen, Zhang, Huang, Shi, Zhang, Yan, Wang, Wang, Liu, and Lian]{shen2025dast}
Shen, Y., Zhang, J., Huang, J., Shi, S., Zhang, W., Yan, J., Wang, N., Wang, K., Liu, Z., and Lian, S.
\newblock Dast: Difficulty-adaptive slow-thinking for large reasoning models, 2026.
\newblock URL \url{https://arxiv.org/abs/2503.04472}.

\bibitem[Shu et~al.(2019)Shu, Chen, Kumar, Ermon, and Poole]{shu2019weakly}
Shu, R., Chen, Y., Kumar, A., Ermon, S., and Poole, B.
\newblock Weakly supervised disentanglement with guarantees.
\newblock \emph{arXiv preprint arXiv:1910.09772}, 2019.

\bibitem[Snell et~al.(2024)Snell, Lee, Xu, and Kumar]{snell2024scaling}
Snell, C., Lee, J., Xu, K., and Kumar, A.
\newblock Scaling llm test-time compute optimally can be more effective than scaling model parameters.
\newblock \emph{arXiv preprint arXiv:2408.03314}, 2024.

\bibitem[Sui et~al.(2025)Sui, Chuang, Wang, Zhang, Zhang, Yuan, Liu, Wen, Zhong, Zou, Chen, and Hu]{sui2025stop}
Sui, Y., Chuang, Y.-N., Wang, G., Zhang, J., Zhang, T., Yuan, J., Liu, H., Wen, A., Zhong, S., Zou, N., Chen, H., and Hu, X.
\newblock Stop overthinking: A survey on efficient reasoning for large language models, 2025.
\newblock URL \url{https://arxiv.org/abs/2503.16419}.

\bibitem[Surana et~al.(2025)Surana, Wu, Xie, Xia, Steck, Liang, Kallus, and McAuley]{surana2025reviews}
Surana, R., Wu, J., Xie, Z., Xia, Y., Steck, H., Liang, D., Kallus, N., and McAuley, J.
\newblock From reviews to dialogues: Active synthesis for zero-shot llm-based conversational recommender system.
\newblock \emph{arXiv preprint arXiv:2504.15476}, 2025.

\bibitem[Talmor et~al.(2019)Talmor, Herzig, Lourie, and Berant]{talmor2019commonsenseqa}
Talmor, A., Herzig, J., Lourie, N., and Berant, J.
\newblock Commonsenseqa: A question answering challenge targeting commonsense knowledge.
\newblock In \emph{Proceedings of the 2019 Conference of the North American Chapter of the Association for Computational Linguistics: Human Language Technologies, Volume 1 (Long and Short Papers)}, pp.\  4149--4158, 2019.

\bibitem[Tan et~al.(2025)Tan, Pan, Lin, Chen, Zheng, Tang, and Yang]{tan2025gtpo}
Tan, H., Pan, J., Lin, J., Chen, T., Zheng, Z., Tang, Z., and Yang, H.
\newblock Gtpo and grpo-s: Token and sequence-level reward shaping with policy entropy.
\newblock \emph{arXiv preprint arXiv:2508.04349}, 2025.

\bibitem[Uesato et~al.(2022)Uesato, Kushman, Kumar, Song, Siegel, Wang, Creswell, Irving, and Higgins]{uesato2022solving}
Uesato, J., Kushman, N., Kumar, R., Song, F., Siegel, N., Wang, L., Creswell, A., Irving, G., and Higgins, I.
\newblock Solving math word problems with process-and outcome-based feedback.
\newblock \emph{arXiv preprint arXiv:2211.14275}, 2022.

\bibitem[Wang et~al.(2026)Wang, Li, Zhang, Wu, Huang, Yao, McAuley, and Shang]{wang2026scenealign}
Wang, C., Li, X., Zhang, J.~Y., Wu, J., Huang, C., Yao, L., McAuley, J., and Shang, J.
\newblock Scenealign: Aligning multimodal reasoning to scene graphs in complex visual scenes.
\newblock \emph{arXiv preprint arXiv:2601.05600}, 2026.

\bibitem[Wang et~al.(2025)Wang, Yu, Wu, Liu, McAuley, and Yao]{wang2025weakly}
Wang, R., Yu, T., Wu, J., Liu, Y., McAuley, J., and Yao, L.
\newblock Weakly-supervised vlm-guided partial contrastive learning for visual language navigation.
\newblock \emph{arXiv preprint arXiv:2506.15757}, 2025.

\bibitem[Wang et~al.(2022)Wang, Wei, Schuurmans, Le, Chi, Narang, Chowdhery, and Zhou]{wang2022self}
Wang, X., Wei, J., Schuurmans, D., Le, Q., Chi, E., Narang, S., Chowdhery, A., and Zhou, D.
\newblock Self-consistency improves chain of thought reasoning in language models.
\newblock \emph{arXiv preprint arXiv:2203.11171}, 2022.

\bibitem[Wei et~al.(2022)Wei, Wang, Schuurmans, Bosma, Xia, Chi, Le, Zhou, et~al.]{wei2022chain}
Wei, J., Wang, X., Schuurmans, D., Bosma, M., Xia, F., Chi, E., Le, Q.~V., Zhou, D., et~al.
\newblock Chain-of-thought prompting elicits reasoning in large language models.
\newblock \emph{Advances in neural information processing systems}, 35:\penalty0 24824--24837, 2022.

\bibitem[Wu et~al.(2024{\natexlab{a}})Wu, Li, Wang, Xia, Xiong, Wang, Yu, Chen, Kveton, Yao, et~al.]{wu2024ocean}
Wu, J., Li, X., Wang, R., Xia, Y., Xiong, Y., Wang, J., Yu, T., Chen, X., Kveton, B., Yao, L., et~al.
\newblock Ocean: Offline chain-of-thought evaluation and alignment in large language models.
\newblock \emph{arXiv preprint arXiv:2410.23703}, 2024{\natexlab{a}}.

\bibitem[Wu et~al.(2024{\natexlab{b}})Wu, Yu, Chen, Wang, Rossi, Kim, Rao, and McAuley]{wu2024decot}
Wu, J., Yu, T., Chen, X., Wang, H., Rossi, R., Kim, S., Rao, A., and McAuley, J.
\newblock Decot: Debiasing chain-of-thought for knowledge-intensive tasks in large language models via causal intervention.
\newblock In \emph{Proceedings of the 62nd Annual Meeting of the Association for Computational Linguistics (Volume 1: Long Papers)}, pp.\  14073--14087, 2024{\natexlab{b}}.

\bibitem[Wu et~al.(2025{\natexlab{a}})Wu, Surana, Xie, Shen, Xia, Yu, Rossi, Ammanabrolu, and McAuley]{wu2025context}
Wu, J., Surana, R., Xie, Z., Shen, Y., Xia, Y., Yu, T., Rossi, R.~A., Ammanabrolu, P., and McAuley, J.
\newblock In-context ranking preference optimization.
\newblock \emph{arXiv preprint arXiv:2504.15477}, 2025{\natexlab{a}}.

\bibitem[Wu et~al.(2025{\natexlab{b}})Wu, Xiong, Li, Hu, Yu, Wang, Chen, Shang, and McAuley]{wu2025ctrls}
Wu, J., Xiong, Y., Li, X., Hu, Z., Yu, T., Wang, R., Chen, X., Shang, J., and McAuley, J.
\newblock Ctrls: Chain-of-thought reasoning via latent state-transition.
\newblock \emph{arXiv preprint arXiv:2507.08182}, 2025{\natexlab{b}}.

\bibitem[Yang et~al.(2025{\natexlab{a}})Yang, Li, Yang, Zhang, Hui, Zheng, Yu, Gao, Huang, Lv, et~al.]{yang2025qwen3}
Yang, A., Li, A., Yang, B., Zhang, B., Hui, B., Zheng, B., Yu, B., Gao, C., Huang, C., Lv, C., et~al.
\newblock Qwen3 technical report.
\newblock \emph{arXiv preprint arXiv:2505.09388}, 2025{\natexlab{a}}.

\bibitem[Yang et~al.(2025{\natexlab{b}})Yang, Yu, Cui, and Wang]{yang2025reasonflux}
Yang, L., Yu, Z., Cui, B., and Wang, M.
\newblock Reasonflux: Hierarchical llm reasoning via scaling thought templates.
\newblock \emph{arXiv preprint arXiv:2502.06772}, 2025{\natexlab{b}}.

\bibitem[Yu et~al.(2025)Yu, Xiong, Wu, Li, Yu, Chen, Sinha, Shang, and McAuley]{yu2025explainable}
Yu, S., Xiong, Y., Wu, J., Li, X., Yu, T., Chen, X., Sinha, R., Shang, J., and McAuley, J.
\newblock Explainable chain-of-thought reasoning: An empirical analysis on state-aware reasoning dynamics.
\newblock \emph{arXiv preprint arXiv:2509.00190}, 2025.

\bibitem[Yuan et~al.(2024)Yuan, Li, Chen, Cui, Ding, Zhang, Zhou, Liu, and Peng]{yuan2024free}
Yuan, L., Li, W., Chen, H., Cui, G., Ding, N., Zhang, K., Zhou, B., Liu, Z., and Peng, H.
\newblock Free process rewards without process labels.
\newblock \emph{arXiv preprint arXiv:2412.01981}, 2024.

\bibitem[Zelikman et~al.(2022)Zelikman, Wu, Mu, and Goodman]{zelikman2022star}
Zelikman, E., Wu, Y., Mu, J., and Goodman, N.
\newblock Star: Bootstrapping reasoning with reasoning.
\newblock \emph{Advances in Neural Information Processing Systems}, 35:\penalty0 15476--15488, 2022.

\bibitem[Zelikman et~al.(2024)Zelikman, Harik, Shao, Jayasiri, Haber, and Goodman]{zelikman2024quiet}
Zelikman, E., Harik, G., Shao, Y., Jayasiri, V., Haber, N., and Goodman, N.~D.
\newblock Quiet-star: Language models can teach themselves to think before speaking.
\newblock \emph{arXiv preprint arXiv:2403.09629}, 2024.

\bibitem[Zhang et~al.(2024)Zhang, Cai, Li, Roberts, Guha, and Sala]{zhang2024stronger}
Zhang, T., Cai, L., Li, J., Roberts, N., Guha, N., and Sala, F.
\newblock Stronger than you think: Benchmarking weak supervision on realistic tasks.
\newblock \emph{Advances in Neural Information Processing Systems}, 37:\penalty0 122292--122315, 2024.

\bibitem[Zhou(2018)]{zhou2018brief}
Zhou, Z.-H.
\newblock A brief introduction to weakly supervised learning.
\newblock \emph{National science review}, 5\penalty0 (1):\penalty0 44--53, 2018.

\end{thebibliography}
